\documentclass{article}
\usepackage{ijcai23}

\usepackage{times}
\usepackage{soul}
\usepackage{url}
\usepackage[hidelinks]{hyperref}
\usepackage[utf8]{inputenc}
\usepackage[small]{caption}
\usepackage{graphicx}
\usepackage{amsmath}
\usepackage{amsthm}
\usepackage{booktabs}
\usepackage{algorithm}
\usepackage{algorithmic}
\usepackage[switch]{lineno}

\usepackage{subcaption} 

\usepackage{arydshln}   %   for \hdashline
\usepackage{graphicx}
\usepackage{bbding}     %   for \XSolidBrush
\usepackage{enumitem}
\usepackage{multicol}
\usepackage{multirow}
\usepackage{caption}
\usepackage{nicefrac}
\usepackage{wrapfig}  % for wrapfigure

\def\eqref#1{equation~\ref{#1}}
\def\Eqref#1{Equation~\ref{#1}}

\newtheorem{lemma}{Lemma}
\newtheorem{property}{Property}

\usepackage{float}
\usepackage{amssymb}
\providecommand*\EnableMain[1]{}
\let\EnableMainstart=\iftrue       % open main
\let\EnableMainend=\fi

\providecommand*\EnableAppx[1]{}
\let\EnableAppxstart=\iftrue       %open appx
\let\EnableAppxend=\fi

\EnableMainstart
    \title{Measuring Acoustics with Collaborative Multiple Agents\footnote{The full paper with appendix together with source code can be found at \url{https://yyf17.github.io/MACMA}. 
    }}
\else
    \EnableAppxstart
    \title{Supplementary Material: \\ Measuring Acoustics with Collaborative Multiple Agents}
    \EnableAppxend
\EnableMainend

\author{
Yinfeng Yu$^{1,6}$
\and
Changan Chen$^2$\and
Lele Cao$^{1,3}$\and
Fangkai Yang$^4$\And
Fuchun Sun$^{1,5,}$\thanks{Corresponding author: Fuchun Sun.}
\affiliations
$^1$Beijing National Research Center for Information Science and Technology, State Key Lab on Intelligent Technology and Systems, Department of Computer Science and Technology, Tsinghua University \\
$^2$University of Texas at Austin \\
$^3$Motherbrain, EQT Group \\
$^4$Microsoft Research \\
$^5$THU-Bosch JCML Center \\
$^6$College of Information Science and Engineering, Xinjiang University
\emails
yyf17@mails.tsinghua.edu.cn,
\hspace{2mm}
changan@cs.utexas.edu,
\hspace{2mm}
lele.cao@eqtpartners.com,
fangkai.yang@microsoft.com,
\hspace{2mm}
fcsun@mail.tsinghua.edu.cn
}

\begin{document}

\maketitle

\EnableMainstart % abstract
\begin{abstract}
As humans, we hear sound every second of our life. The sound we hear is often affected by the acoustics of the environment surrounding us. For example, a spacious hall leads to more reverberation. Room Impulse Responses (RIR) are commonly used to characterize environment acoustics as a function of the scene geometry, materials, and source/receiver locations. Traditionally, RIRs are measured by setting up a loudspeaker and microphone in the environment for all source/receiver locations, which is time-consuming and inefficient. We propose to let two robots measure the environment's acoustics by actively moving and emitting/receiving sweep signals. We also devise a collaborative multi-agent policy where these two robots are trained to explore the environment's acoustics while being rewarded for wide exploration and accurate prediction. We show that the robots learn to collaborate and move to explore environment acoustics while minimizing the prediction error. To the best of our knowledge, we present the very first problem formulation and solution to the task of collaborative environment acoustics measurements with multiple agents.
\end{abstract}
\EnableMainend

%####################################################
%%%%%%%%%%%%%%%%%%%%%%%%%%%%%%%%%%%%%%%%%%%%%%%%%%%%%%%%%%%%%%%%%%%%%%%%
\EnableMainstart  % start of main
%%%%%%%%%%%%%%%%%%%%%%%%%%%%%%%%%%%%%%%%%%%%%%%%%%%%%%%%%%%%%%%%%%%%%%%%

% \section{INTRODUCTION}
\section{Introduction}

\begin{figure}[!th]
% \vspace{-20pt}
\centering
    \includegraphics[width=0.48\textwidth]{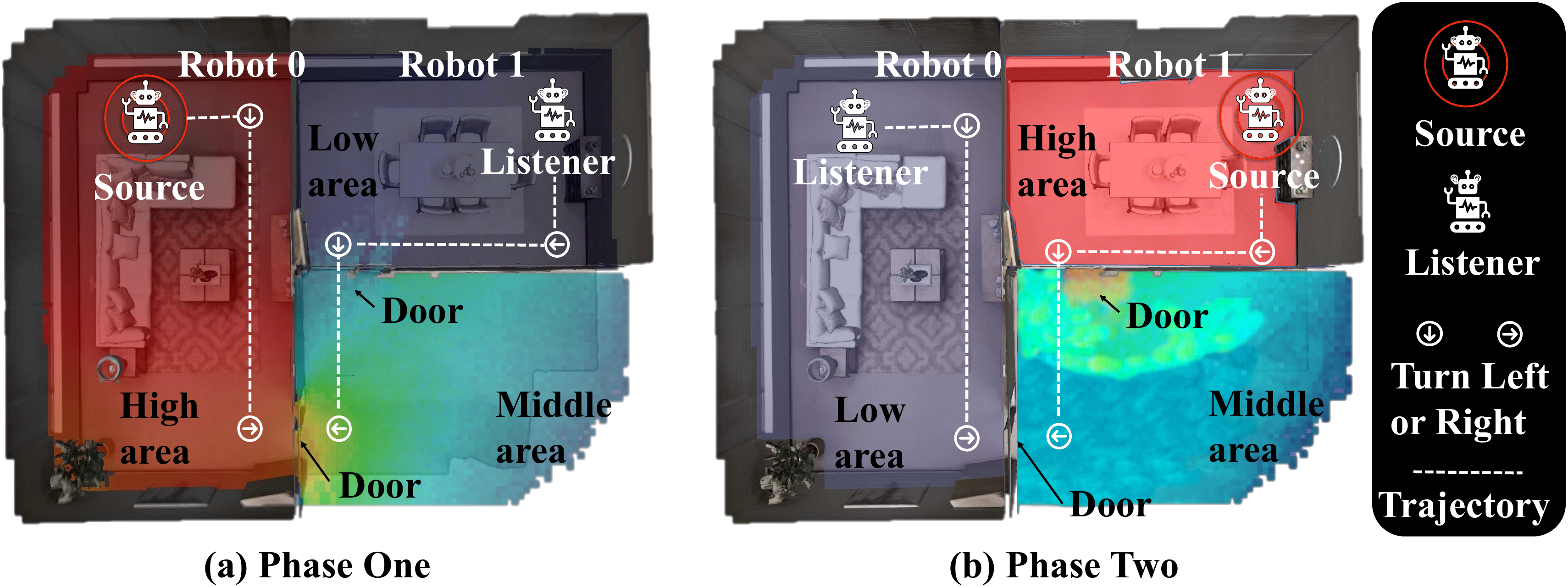}
%   \vspace{-8pt}
  \caption{\small Learn to measure environment acoustics with two collaborative robots. 
The background color indicates sound intensity (``High'', ``Middle'' and ``Low'' areas).
Each step (one step per second) embodies three steps: 1) robot 0 emits a sound, and robot 1 receives the sound; 2) robot 1 emits the sound, and robot 0 receives the sound; 3) two robots make a movement following their learned policies. 
This process repeats until reaching the maximum number of time steps.}
\label{fig: title-fig}
  \vspace{-6mm}
\end{figure}

Sound is critical for humans to perceive and interact with the environment.
Before reaching our ears, sound travels via different physical transformations in space, such as reflection, transmission and diffraction. 
These transformations are characterized and measured by a Room Impulse Response (RIR) function~\cite{valimaki2016more}.
RIR is the transfer function between the sound source and the listener (microphone).
Convolving the anechoic sound with RIR will get the sound with reverberation~\cite{Manocha16}.
RIR is utilized in many applications such as sound rendering~\cite{schissler2014high}, sound source localization~\cite{TangCWYM20}, audio-visual matching~\cite{VAM}, and audio-visual navigation~\cite{SoundSpaces,Wan-AVN,SAVN,SAAVN}.  
For example, to achieve clear speech in a concert hall, one might call for a sound rendering that drives more acoustic reverberation while keeping auditoriums with fewer reverberation \cite{MildenhallSTBRN22}.
The key is to measure RIR at different locations in the hall. 
However, RIR measuring is time-consuming due to the large number of samples to traverse.
To illustrate, in a $5\!\!\times\!\!5$ m$^2$ room with a spatial resolution of 0.5m,  the number of measurable points is $11\!\!\times\!\!11$=121.
The source location (omnidirectional) can sample one of these 121 points.
Assuming a listener with four orientations (0, 90, 180, 270), this listener can choose from 121 points with four directions for each chosen point.
So, the number of source-listener pairs becomes $121\!\times\!121\!\times\!4$=58,564.
Assuming the sampling rate, duration and precision of binaural RIR is 16K, 1 second and float32 respectively, one RIR sample requires $2\!\times\!16000 \times\!4$ Bytes = 128KB from computer storage (memory).
The entire room would take up to $58,564\!\times\!128$ KB $\approx\!7.5$ GB.
Moreover, it also means that one has to move the source/listener devices 58,564 times and performs data sending/receiving for each point.

There are some attempts to solve the challenge of storage: 
FAST-RIR~\cite{fast-rir} relies on handcrafted features, source (emitter) and listener (receiver) locations to generate RIR while being fidelity agnostic; 
MESH2IR~\cite{ratnarajah2022mesh2ir} uses the scene's mesh and source/listener locations to generate RIR while ignoring the measurement cost; 
Neural Acoustic Field (NAF)~\cite{naf} tries to learn the parameters of the acoustic field, but its training time and model storage cost grow linearly with the number of environments~\cite{FEW-SHOTRIR}.
Some work~\cite{Image2Reverb} suggests that storing the original RIR data of the sampled points is optional, and only the acoustic field parameters must be stored.
However, given a limited number of action steps, it is challenging to model the acoustic field.

To overcome the aforementioned challenges, we propose MACMA (Measuring Acoustics with Collaborative Multiple Agents) which is illustrated in Figure~\ref{fig: title-fig}).
Both agents, one source (emitter) and one listener (receiver), learn a motion policy to perform significance sampling of RIR within any given 3D scene.
The trained agents can move (according to the learned motion policy) in any new 3D scene to predict the RIR of that new scene.
To achieve that, we design two policy learning modules: the RIR prediction module and the dynamic allocation module of environment reward.
\EnableAppxstart
In Appx.~\ref{appx: m2mrirra},
\else 
In Appx. B,
\EnableAppxend we explore the design of environmental reward and based on this, and we further propose a reward distribution module to learn how to efficiently distribute the reward obtained at the current step, thereby incentivizing the two agents to learn to cooperate and move.
To facilitate the convergence of optimization, we design loss functions separately for the policy learning module, the RIR prediction module, and the reward allocation module.
Comparative experiments and ablation experiments are performed on two datasets Replica~\cite{replica} and Matterport3D~\cite{matterport3d}, verifying the effectiveness of the proposed solution.
To the best of our knowledge, this work is the first RIR measurement method using two collaborative agents.
The main contributions of this work are:
\begin{itemize}
     \item we propose a new setting for planning RIR measuring under finite time steps and a solution to measure the RIR with two-agent cooperation in low resource situations;
     \item we design a novel reward function for the multi-agent decomposition to encourage coverage of environment acoustics;
     \item we design evaluation metrics for the collaborative measurement of RIR, and we experimentally verify the effectiveness of our model. 
\end{itemize}

%-------------------------------------------------
\begin{figure*}[t!]
    \centering
    \includegraphics[width=0.86\textwidth]{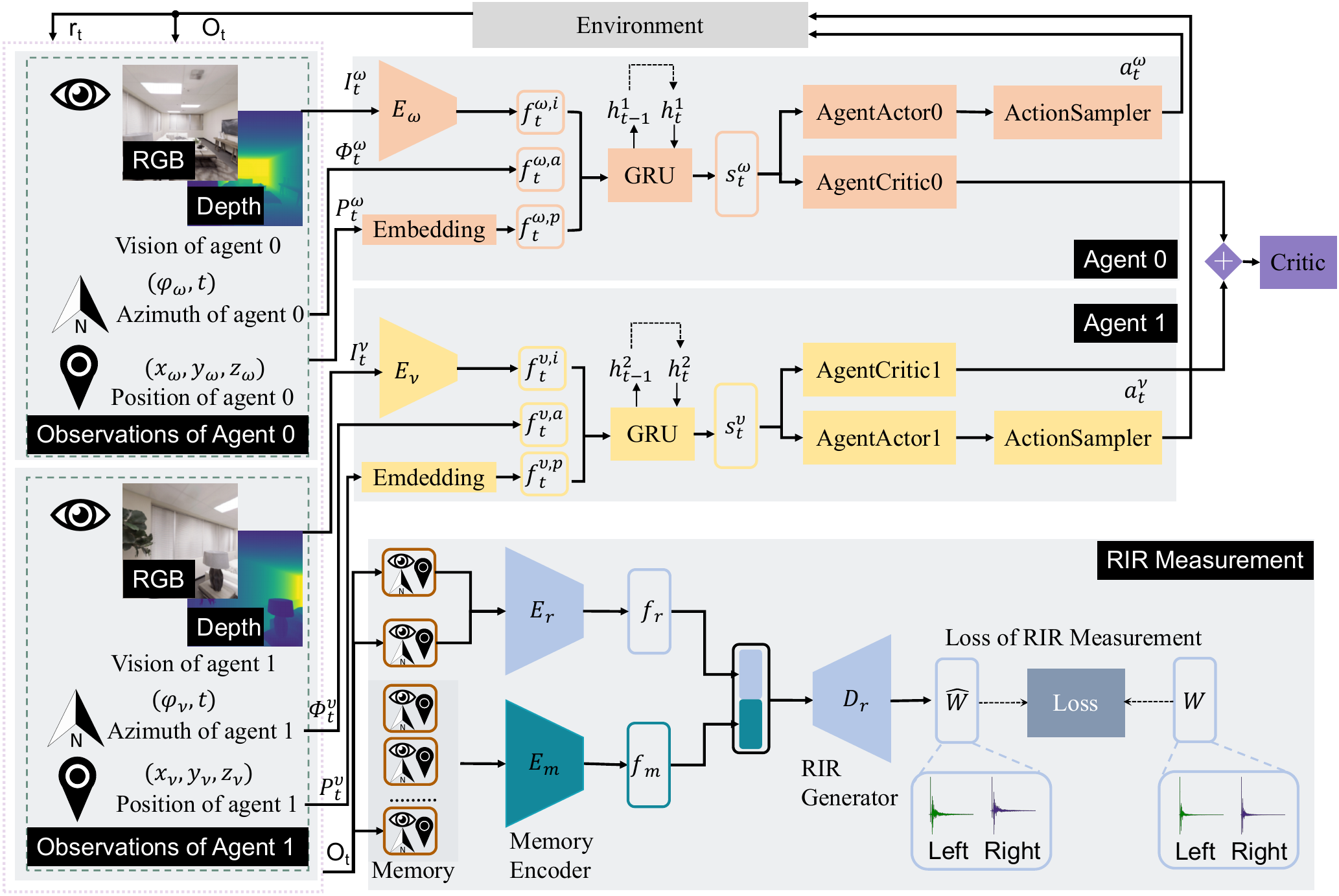}
    \caption{
        The MACMA architecture: the agent 0 and the agent 1 first learn to encode observations as $s^{\omega}_{t}$ and $s^{\nu}_{t}$ respectively using encoder $E_{\omega}$ and $E_{\nu}$, which are fed to actor-critic networks to predict the next action $a_t^{\omega}$ and $a_t^{\nu}$.
        The RIR Measurement learns how to predict room impulse response $\hat{W}_{t}$ guided by ground truth $W_{t}$. 
    }
    \label{fig: network}
\end{figure*}
%----------------------------------------------------------------------------------------

% \section{RELATED WORK} \label{related-work}
\section{Related Work} \label{related-work}

\noindent{\textbf{RIR generation. }}
Measuring the RIR has been of long-standing interest to researchers~\cite{Manocha16,savioja2015overview}. 
Traditional methods for generating RIR include statistical based methods~\cite{schissler2014high,ZhenyuTangARM22} and physics-based methods~\cite{MehraRACCM13,TaylorCMLSM12}.
However, they are computationally prohibitive. Recent methods estimate the acoustics of RIR by parameters to generate RIR indirectly~\cite{StoRIR,gpuRIR,IR-GAN}.
Although these methods are flexible in extracting different acoustic cues, their predictions are independent of the source and receiver's exact locations, making them unsuitable for scenarios where the mapping between RIR and locations of source and receiver is important (e.g. sound source localization and audiovisual navigation).
FAST-RIR~\cite{fast-rir} is a GAN-based RIR generator that generates a large-scale RIR dataset capable of accurately modeling source and receiver locations, but for efficiency, they rely on handcrafted features instead of learning them, which affects generative fidelity~\cite{naf}.
Neural Acoustic Field (NAF)~\cite{naf} addresses the issues of efficiency and fidelity by learning an implicit representation of RIR~\cite{MildenhallSTBRN22}, and by introducing global features and local embeddings.
However, NAF cannot generalize to new environments, and its training time and model storage cost grows linearly with the number of environments~\cite{FEW-SHOTRIR}.
The recently proposed MESH2IR~\cite{ratnarajah2022mesh2ir} is an indoor 3D scene IR generator that takes the scene's mesh, listener positions, and source locations as input.
MESH2IR~\cite{ratnarajah2022mesh2ir} and FAST-RIR~\cite{fast-rir} assume that the environment and reverberation characteristics have been given, hence they only consider the fitting for the existing dataset, and ignore the measurement cost.
However, our model considers how two moving agents collaborate to optimize RIR measuring from the aspects of time consumption, coverage, accuracy, etc. 
It is worth mentioning that our model addresses multiple scenarios, so that it is more suitable for generalizing to unseen environments.

\noindent{\textbf{Audio spatialization. }}
Binaural audio generation methods comprise converting mono audio to binaural audio using visual information in video~\cite{Multi-Task-Binaural}, utilizing spherical harmonics to generate binaural audio from mono audio for training~\cite{XuZ0D0L21}, and generate binaural audio from video~\cite{gao20192}.
Using 360 videos from YouTube to generate 360-degree ambisonic sound~\cite{morgado2018self} is a higher-dimensional audio spatialization.
Alternatively, \cite{RachavarapuAS021} directly synthesize spatial audio.
Audio spatialization has a wide range of practical applications, such as object/speaker localization~\cite{av-Localization}, speech enhancement~\cite{MichelsantiTZXY21}, speech recognition~\cite{ShaoZY22}, etc.
Although all the above works use deep learning, our work is fundamentally different in that we propose to model binaural RIR measuring as a decision process (of two moving agents that learns to plan measurement) using time-series states as input.

\noindent{\textbf{Audio-visual learning. }}
\cite{FEW-SHOTRIR} harnesses the synergy of egocentric visual and echogenic responses to infer ambient acoustics to predict RIR.
The advancement of audiovisual learning has good applications in many tasks, such as audiovisual matching~\cite{VAM}, audiovisual source separation~\cite{av-Separation} and audiovisual navigation~\cite{SoundSpaces,SHE,LLA,fallen-obj-avn,fsaavn,E3VN}.
There are also work to use echo responses with vision to learn better spatial representations~\cite{VisualEchoes}, infer depth~\cite{BatVision}, or predict floor plans of 3D environments~\cite{PurushwalkamGIS21}.
The closest work to ours is audio-visual navigation, but audio-visual navigation has navigation goals, but the agents in our setup have no clear navigation destinations.

\noindent{\textbf{Multi-agent learning. }}
% There are two types of collaborative multi-agents: value decomposition based methods~\cite{mavdn,maqmix,qtran,maven,feudalma,asn,qpd,rode,ndq,roma,edti,ciexplore,wqmix,lica,CollaQ,QPLEX,Qatten,masoftQ} and actor-critic~\cite{COMA,dop,actorattencritic,MADDPG} based methods.
There are two types of collaborative multi-agents: value decomposition based methods~\cite{maqmix,wqmix} and actor-critic~\cite{COMA,MADDPG} based methods.
The centralized training with decentralized execution (CTDE) paradigm~\cite{dop} has recently attracted attention for its ability to address non-stationarity while maintaining decentralized execution. 
Learning a centralized critic with decentralized actors (CCDA) is an efficient approach that exploits the CTDE paradigm.
Multi-agent deep deterministic policy gradient (MADDPG) and counterfactual multi-agent (COMA) are two representative examples. 
In our design, we have a centralized critic (named Critic) with decentralized actors (named AgentActor0 and AgentActor1).
But our task differs from all the above multi-agent learning methods in that the agents in our scenario are working on a collaborative task, while previous multi-agent learning research has mainly focused on competitive tasks.

% \section{THE PROPOSED APPROACH} \label{sec: approach}
\section{The Proposed Approach} \label{sec: approach}

Our model MACMA has two collaborative agents moving in a 3D environment in Figure~\ref{fig: title-fig}, using vision, position, and azimuth to measure the RIR.
The proposed model mainly consists of three parts: agent 0, agent 1, and RIR measurement (see Figure~\ref{fig: network}).
Given egocentric vision, azimuth, and position inputs, our model encodes these multi-modal cues to 1) determine the action for agents and evaluate the action taken by the agents for policy optimization, 2) measure the room impulse response and evaluate the regression accuracy for the RIR generator, and 3) evaluate the trade-off between the agents' exploration and RIR measurement.
The two agents repeat this process until the maximal steps have been reached.

Specifically, at each step $t$ (cf. Figure~\ref{fig: network}), the robots receive the current observation of their own $O^{\omega}_{t}$ and $O^{\nu}_{t}$ respectively, 
where $O^{\omega}_{t}=(I^{\omega}_{t},\,\Phi^{\omega}_{t},\,P^{\omega}_{t} )$, $O^{\nu}_{t}=(I^{\nu}_{t},\,\Phi^{\nu}_{t},\,P^{\nu}_{t} )$,
$I^{\omega}_{t} = (I^{\omega,\,rgb}_{t}, I^{\omega,\,depth}_{ t}) $ and $\,I^{\nu}_{t} = (I^{\nu,\,rgb}_{t}, I^{\nu,\,depth}_{t})$ are egocentric visions for robot 0 and robot 1 respectively. 
$\Phi^{\omega}_{t}=\left(\varphi^{\omega}_{t}, t\right)\,$ and $\Phi^{\nu}_{t}=\left(\varphi_{\nu}, t\right)$ are azimuths for robot 0 and robot 1 respectively. 
$P^{\omega}_{t}=\left(x_{\omega}, y_{\omega}, z_{\omega}\right)\,$ and $P^{\nu}_{t}=\left(x_{v}, y_{v}, z_ { v}\right)$ are positions for robot 0 and robot 1 respectively. 
$I^{\omega}_{t}$ or $I^{\nu}_{t}$ denotes the current visual input that can be RGB (128$\times$128$\times$3 pixels) and/or depth (with a dimension of 128$\times$128$\times$1) image\footnote{Both RGB and depth images capture the 90-degree field of view in front of the navigating robot.}, 
$\Phi^{\omega}_{t}$ and $\Phi^{\nu}_{t}$ are 2D vector with time.
$P^{\omega}_{t}$ and $P^{\nu}_{t}$ are 3D vector. 
Although there exists a {\it navigability graph} (with nodes and edges) of the environment,
this graph is hidden from the robot, hence the robots must learn from the accumulated observations $O^{\omega}_{t}$ and $O^{\nu}_{t}$ to understand the geometry of the scene.
At each step, the robot at a certain node A can only move to another node B in the navigability graph if 1) an edge connects both nodes, and 2) the robot is facing node B.
The viable robotic action space is defined as $\mathcal{A}$ =\{\verb|MoveForward|, \verb|TurnLeft|, \verb|TurnRight|, \verb|Stop|\}, where the \verb|Stop| action should be executed when the robot completes the task or the number of the robot's actions reach the maximum number of steps.
The overall goal is to {\bf predict RIR in new scene accurately and explore widely}.
%----------------------------------------------------------------------------------------

\subsection{Problem Formulation} \label{main: problem}

We denote agent 0 and agent 1 with superscript $\omega$ and $\nu$, respectively. 
The game $\mathcal{M} = (\mathcal{S}, (\mathcal{A}^{\omega}, \mathcal{A}^{\nu}) , \mathcal{P},(\mathcal{R}^{\omega}, \mathcal{R}^{\nu}))$ consists of state set $\mathcal{S}$, action sets ($\mathcal{A}^{\omega}\,$, $\mathcal{A}^{\nu}\,$), a joint state transition function $\mathcal{P}\,$: $\mathcal{S} \times \mathcal{A}^{\omega} \times \mathcal{A}^{\nu} \rightarrow \mathcal{S}$,
and the reward functions $\mathcal{R}^{\omega}\,: \mathcal{S} \times \mathcal{A}^{\omega} \times \mathcal{A}^{\nu} \times \mathcal{S} \rightarrow \mathbb{R} \,$ for agent 0 and $\mathcal{R}^{\nu} : \mathcal{S} \times \mathcal{A}^{\omega} \times \mathcal{A}^{\nu} \times \mathcal{S} \rightarrow \mathbb{R} \,$ for agent 1.
Each player wishes to maximize their discounted sum of rewards. 
$r$ is the reward given by the environment at every time step in an episode.
MACMA is modeled as a multi-agent~\cite{mavdn,maqmix} problem involving two collaborating players sharing the same goal:
\begin{equation} \label{eq: total-ques}
    \small
    \begin{aligned}
        \text{min.} & \; \mathcal{L} \quad
        \text{s.t.} \; \pi^{\star} = (\pi^{\star,\,\omega},\pi^{\star,\,\nu}) = \!\!\!\!
        \mathop{\arg\max}\limits_{\pi^{\omega} \in \Pi^{\omega},\,\pi^{\nu} \in \Pi^{\nu} } G(\pi^{\omega},\,\pi^{\nu},\,r) \\
        \text{where }& \quad G(\pi^{\omega}\!,\pi^{\nu}\!,r)\!=\! w^{\omega}G(\pi^{\omega}\!,r) + w^{\nu}G(\pi^{\nu}\!,r), \\
        & \; G(\pi^{\omega}\!,r)\!=\!\textstyle\sum_{t=0}^{T-1} \gamma^{t}r_{t}\rho^{\omega},
        \quad G(\pi^{\nu}\!,r)\!=\!\sum_{t=0}^{T-1} \gamma^{t}r_{t}\rho^{\nu},\\
        \quad & \quad  \rho^{\omega}\!\!= (1-\rho)/2,  
        \quad \rho^{\nu}\!\!= (1-\rho)/2,  \\
        \quad &  \quad  w^{\omega} > 0, \quad w^{\nu} > 0, \quad 0 \le \rho \le 1 \quad \text{or} \quad \rho = -1.0, \quad \\
    \end{aligned}
\end{equation}
where the loss $\mathcal{L}$ is defined in~\Eqref{eq: loss-total}.
$G(\pi^{\omega},\,\pi^{\nu},\,r)$ is the expected joint rewards for agent 0 and agent 1 as a whole. 
$G(\pi^{\omega},\,r)$ and $G(\pi^{\nu},\,r)$ are the discounted and cumulative rewards for agent 0 and agent 1, respectively.
$w^{\omega}$ and $w^{\nu}$ denote the constant cumulative rewards balance factors for agent 0 and agent 1, respectively.
$\rho^{\omega}$ and $\rho^{\nu}$ are immediate reward contributions for agent 0 and agent 1, respectively.
$\rho$ is a constant (throughout the training) reward allocation parameter. 
Inspired by \textit{Value Decomposition Networks} (VDNs)~\cite{mavdn} and QMIX~\cite{maqmix}, we construct the objective function $G(\pi^{\omega},\pi^{\nu},r)$ in~\Eqref{eq: total-ques} by combining the non-negative partial reciprocal constraint respond to $G(\pi^{\omega},r)$ and $G(\pi^{\nu},r)$ \EnableAppxstart(see theoretical details in Appx.~\ref{appx:theoretical}).
\else
(see theoretical details in Appx. A.1).
\EnableAppxend

\noindent{\textbf{Agent 0, agent 1 and their optimization. }}
The agent 0 and agent 1 receive the current observation $O^{\omega}_{t}=(I^{\omega}_{t},\,\Phi^{\omega}_{t},\,P^{\omega}_{t} )$ and $O^{\nu}_{t}=(I^{\nu}_{t},\,\Phi^{\nu}_{t},\,P^{\nu}_{t} )$ at the $t$-th step.
The visual ($I^{\omega}_{t}$ and $I^{\nu}_{t}$) part is encoded into a visual feature vector using a CNN encoder: $f_{t}^{\omega,\,i}$ and $f_{t}^{\nu,\,i}$ ($E_{\omega}$ for agent 0 and $E_{\nu}$ for agent 1).
Visual CNN encoders $E_{\omega}$ and $E_{\nu}$ are constructed in the same way (from the input to output layer): \verb|Conv8x8|, \verb|Conv4x4|, \verb|Conv3x3| and a 256-dim linear layer; 
ReLU activations are added between any two neighboring layers.
$P^{\omega}_{t}$ and $P^{\nu}_{t}$ are embedded by an embedding layer and encoded into feature vectors $f_{t}^{\omega,\,p}$ and $f_{t}^{\nu,\,p}$, respectively.
Then, we concatenate the two vectors together with $f_{t}^{\omega,\,a}$ ($\Phi^{\omega}_{t}$) and $f_{t}^{\nu,\,a}$ ($\Phi^{\nu}_{t}$) to obtain the global observation embedding $e^{\omega}_{t}=[f_{t}^{\omega,\,i},\,f_{t}^{\omega,\,a},\,f_{t}^{\omega,\,p}]$ and $e^{\nu}_{t}=[f_{t}^{\nu,\,i},\,f_{t}^{\nu,\,a},\,f_{t}^{\nu,\,p}]$.
We transform the observation embeddings to state representations using a gated recurrent unit (GRU), $s^{\omega}_{t} = \normalfont\textsc{GRU}(e^{\omega}_t,\,h^{1}_{t-1})$.
We adopt a similar procedure to obtain  $s^{\nu}_{t} = \normalfont\textsc{GRU}(e^{\nu}_t,\,h^{2}_{t-1})$.
The state vectors ($s^{\omega}_{t}$ for agent 0 and $s^{\nu}_{t}$ for agent 1) are then fed to an actor-critic network to 1) predict the conditioned action probability distribution $\pi_{\theta_{1}^{\omega}}(a^{\omega}_t|s^{\omega}_{t})$ for agent 0 and $\pi_{\theta_{1}^{\nu}}(a^{\nu}_t|s^{\nu}_{t})$ for agent 1, 
and 2) estimate the state value $V_{\theta_{2}^{\omega}}(s^{\omega}_{t},\,r^{\omega}_ {t})$ for agent 0 and $V_{\theta_{2}^{\omega}}(s^{\nu}_{t},\,r^{\nu}_ {t})$ for agent 1.
The actor and critic are implemented with a single linear layer parameterized by $\theta_{1}^{\omega}$, $\theta_{1}^{\nu}$, $\theta_{2}^{\omega}$, and $\theta_{2}^{\nu}$, respectively.
For the sake of conciseness, we use $\boldsymbol\theta$ to denote the compound of $\theta_{1}^{\omega}$, $\theta_{1}^{\nu}$, $\theta_{2}^{\omega}$, and $\theta_{2}^{\nu}$ hereafter.
The action samplers in Figure~\ref{fig: network} sample the actual action (i.e. $a_t^{\omega}$ for agent 0 and $a_t^{\nu}$ for agent 1) to execute from $\pi_{\theta_{1}^{\omega}}(a^{\omega}_t|s^{\omega}_{t})$ for agent 0 and $\pi_{\theta_{1}^{\nu}}(a^{\nu}_t|s^{\nu}_{t})$ for agent 1, respectively. 
Both agent 0 and agent 1 optimize their policy by maximizing the expected cumulative rewards $\normalfont\textsc{G}(\pi^{\omega}, r)$ and $\normalfont\textsc{G}(\pi^{\nu}, r)$ respectively in a discounted form.
The Critic module evaluates the actions taken by agent 0 and agent 1 to guide them to take an improved action at the next time step. 
The loss of $\mathcal{L}^{m}$ is formulated as \Eqref{eq: loss-motion}.
%----------------------------------------------------------------------------------------
\begin{equation} \label{eq: loss-motion}
    \mathcal{L}^{m} = w^{\omega}_{m} \cdot \mathcal{L}^{\omega}_{m} +  w^{\nu}_{m} \cdot \mathcal{L}^{\nu}_{m},
\end{equation}
%----------------------------------------------------------------------------------------
where $\mathcal{L}^{\omega}_{m}$ and $\mathcal{L}^{\nu}_{m}$ are motion loss for agent 0 and agent 1 respectively,
$w^{\omega}_{m}, w^{\nu}_{m}$ are hyperparameters.
The loss $\mathcal{L}^{j}_{m}$ is defined as
%----------------------------------------------------------------------------------------
\begin{equation} \label{eq: loss-motion-j}
    \begin{split}
        \mathcal{L}^{j}_{m} = &\sum 0.5\left(\hat{V}_{\theta^{j}}(s) - V^{j}(s)\right)^{2} \\
        &\!-\!\sum\left[\hat{A}^{j}\!\log(\pi_{\theta^{j}}(a\!\mid\!s))
        + \beta \cdot H(\pi_{\theta^{j}}(a\!\mid\!s))\right], \\
    \end{split}
\end{equation}
%----------------------------------------------------------------------------------------
where $j \in \{\omega,\,\nu\}$, and the estimated state value of the target network for $j$ is denoted as $\hat{V}_{\theta^{j}}(s)$.
$\small\smash{\normalfont\textsc{V}^{j}(s) = \max_{a \in \mathbb{A}^{j}}\mathbb{E}[r_{t}+\gamma \cdot \normalfont\textsc{V}^{j}(s_{t+1}) \mid s_{t} = s]}$. 
The advantage for a given length-$T$ trajectory is: $\small\smash{\hat{\normalfont\textsc{{A}}}_{t}^{j} = \sum_{i=t}^{T-1}\gamma^{i+2-t} \cdot \delta_{i}^{j}}$, where $\delta_{t}^{j} = r_{t}+\gamma \cdot \normalfont\textsc{V}^{j}(s_{t+1})-\normalfont\textsc{V}^{j}(s_{t})$.
$H(\pi_{\theta^{j}}(a \mid s))$ is entropy of $\pi_{\theta^{j}}(a \mid s)$.
We collectively denote all the weights in Figure~\ref{fig: network} except the above actor-critic network for agent 0 and agent 1 as $\boldsymbol\Omega$ hereafter for simplicity.

\noindent{\textbf{RIR measurement and its regression. }}
We encode the observations $O^{\omega}_{t}$ and $O^{\nu}_{t}$ with encoder $E_{r}$, and the output of the encoder $E_{r}$ is $f_{r}$.
The historical observations ${O^{\omega}_{t+1-\kappa},\,O^{\nu}_{t+1-\kappa},\,\,\cdots,\,\,O^{\omega}_{t-1},\,\,O^{\nu}_{t-1},\,\,O^{\omega}_{t},\,\,O^{\nu}_{t}}$ are sorted in the memory, and are encoded by $E_{m}$ outputting $f_{m}$, where $\kappa$ is the length of the memory bank.
Then, $f_{r}$ and $f_{m}$ are concatenated.
The predicted RIR $\hat{W}_{t}$ is obtained using RIR generator $D_{r}$.
\EnableAppxstart
For more details for the structure of $E_{r}$, $E_{m}$ and $D_{r}$, please refer to Appx.~\ref{appx: structure-edr}.
\else
For more details on the structure of $E_{r}$, $E_{m}$ and $D_{r}$, please refer to Appx. A.2.
\EnableAppxend
RIR measurement is learned with the ground truth RIR $W_{t}$.
$\mathcal{L}^{\xi}$ denote the loss of RIR measurement.
$\mathcal{L}^{\xi}$ are formulated as

\vspace{-0.1in}
\begin{equation} \label{eq: loss-rir-prediction}
    % \vspace{-0.2in}
    \begin{aligned}
       \mathcal{L}^{\xi}\!&=\!(1\!-\!w^{\text{MSE}})\!\cdot\!10\!\cdot\!\mathcal{L}^{\text{STFT}}\!+\!w^{\text{MSE}}\!\cdot\!4464.2\!\cdot\!\mathcal{L}^{\text{MSE}},   \\
       \mathcal{L}^{\text{STFT}}\!&=\!\sum\!\Delta(W_{t},\,\hat{W}_{t}),\quad
       \mathcal{L}^{\text{MSE}}\!=\!\sum\!\text{MSE}(W_{t},\,\hat{W}_{t}),
    \end{aligned}
\end{equation}
where $\hat{W}_{t}$ is the predicted RIR from the RIR Measurement module. 
$W_{t}$ is the ground truth RIR.
$\Delta(W_{t},\,\hat{W}_{t})$ is STFT (Short-time Fourier transform) distance.
It is calculated by \Eqref{metric: stft-distance}.
10 and 4464.2 are experimental parameters from grid search.

\noindent{\textbf{The total evaluation. }}
The Critic module is implemented with a linear layer.
The total loss of our model is formulated as \Eqref{eq: loss-total}.
We minimize $\mathcal{L}$ following Proximal Policy Optimization (PPO)~\cite{ppo}.
\begin{equation} \label{eq: loss-total}
    \mathcal{L} = w_{m} \cdot \mathcal{L}^{m} + w_{\xi} \cdot \mathcal{L}^{\xi}  ,
\end{equation}
where $\mathcal{L}^{m}$ is the loss component of motion for two agents,
$\mathcal{L}^{\xi}$ is the loss component of room impulse response prediction,
$w_{m}\,$ and $w_{\xi}\,$ are hyperparameters.
The losses $\mathcal{L}^{m}$ and $\mathcal{L}^{\xi}$ are formulated in \Eqref{eq: loss-motion} and \Eqref{eq: loss-rir-prediction}, respectively.

%---------------------------------
\noindent{\textbf{The design of environmental reward.}} \label{sec: reward}
\begin{figure}[t!]
    % \vspace{-20pt}
      \begin{center}
        \includegraphics[width=0.25\textwidth,scale=0.1]{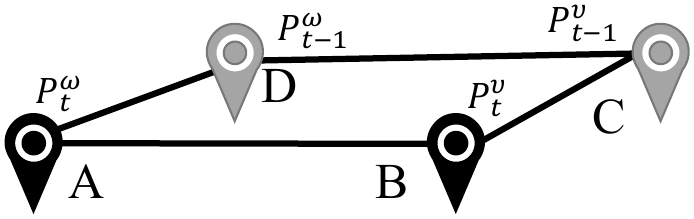}
      \end{center}
      \vspace{-8pt}
      \caption{ \small Demonstration of the current (A and B) and previous (C and D) positions of two robots. The above four coplanar points are denoted as $\Gamma_{ABCD}$.
      }
    \label{fig: convex}
    \vspace{-3mm}
\end{figure}
It can be seen from Figure~\ref{fig: convex} that the $\Gamma_{ABCD}$ is formed by the positions of the two agents at the current time step and the previous time step.
$r_{t}$ is the current step reward, which is calculated by the following \Eqref{eq: r-def}.
\begin{equation} \label{eq: r-def}
    r_{t}= r^{\xi}_{t} + r^{\zeta}_{t} + r^{\psi}_{t} + r^{\phi}_{t},
\end{equation}
%------------------------------------------------------------
$\xi$ denotes the prediction of room impulse response.
$\zeta$ denotes coverage rate.
$\psi$ denotes the length of the perimeter of the convex hull.
$\phi$ denotes the area of the perimeter of the convex hull.

Among them, $r^{\xi}_{t}$ is the reward component in terms of measurement accuracy, which evaluates the improvement of the reward of the measurement accuracy of the current step and the reward of the measurement accuracy of the previous step.
$r^{\xi}_{t}$ is calculated by
%------------------------------------------------------------
\begin{equation} \label{eq: r-rir}
    r^{\xi}_{t} = \alpha^{\xi} \cdot ( \xi_{t} - \xi_{t-1} ),  \quad \xi_{t} = - \Delta(W_{t},\,\hat{W}_{t}),
\end{equation}
%------------------------------------------------------------
where $\xi_{t}$ is the measurement accuracy of the current step.
As briefly explained before, $\Delta(W_{t},\,\hat{W}_{t})$ is the STFT distance that can be calculated by
\begin{equation} \label{metric: stft-distance}
    \Delta(W_{t},\,\hat{W}_{t}) = 0.5 \cdot \Theta(z,\hat{z}) + 0.5 \cdot \Xi(z,\hat{z}),
\end{equation}
where $z$ is the magnitude spectrogram of ground truth RIR $W_{t}$ for the current time step, while 
$\hat{z}$ is the corresponding predicted variant. 
$\Theta(z,\hat{z})$	is the average loss of spectral convergence for $z$ and $\hat{z}$; and $\Xi(z,\hat{z})$ is the log STFT magnitude loss.
$\Theta(z,\hat{z})$ and $\Xi(z,\hat{z})$ are computed with
\begin{equation} \label{metric: stft-distance-theta}
   \Theta(z,\hat{z}) = \frac{\lVert z - \hat{z} \rVert_{F}}{\lVert z \rVert_{F}}
   \quad\text{and}\quad
   \Xi(z,\hat{z}) = \sum \left| \log(\frac{z}{\hat{z}}) \right|,
\end{equation}
where $\lVert \cdot \rVert_{F}$ is Frobenius Norm.
$z=\varLambda(W_{t})=\sqrt{y_{r}^{2}+y_{i}^{2}}$,
where $y_{r}$ is real part of STFT transform\footnote{The parameters for STFT transform are $\#$FFT=1024, $\#$shift =120, $\#$window=600, window=``Hamming window''.} of $W_{t}$,
$y_{i}$ is an imaginary part of the result of the STFT transform of $W_{t}$.
$\hat{z}=\varLambda(\hat{W}_{t})$ is defined similarly to $z$,
and the calculation process of both $z$ and $\hat{z}$ are the same.
	
%------------------------------------------------------------
${\zeta}_{t}$ is the coverage of the current step, which is the ratio of visited nodes (only one duplicate node is counted) to all nodes in the scene at time step $t$.
We calculate $r^{\zeta}_{t}$ by
%------------------------------------------------------------
\begin{equation} \label{eq: r-cr}
    r^{\zeta}_{t} = \alpha^{\zeta} \cdot ( {\zeta}_{t} - {\zeta}_{t-1} ) . 
\end{equation}
%------------------------------------------------------------
$\psi_{t}$ and $\phi_{t}$ are respectively the perimeter and area of $\Gamma_{ABCD}$ in Figure~\ref{fig: convex} at time step $t$.
We calculate $r^{\psi}_{t}$ and $r^{\phi}_{t}$ with
%------------------------------------------------------------
\begin{equation} \label{eq: r-cl}
    r^{\psi}_{t} = \alpha^{\psi} \cdot ( \psi_{t} - \psi_{t-1}) 
    \quad \text{and} \quad
    r^{\phi}_{t} = \alpha^{\phi} \cdot ( \phi_{t} - \phi_{t-1} )  ,
\end{equation}
where $\alpha^{\xi}=1.0$, $\alpha^{\zeta}=1.0$, $\alpha^{\psi}=-1.0$ and $\alpha^{\phi}=1.0$ are hyperparameters \EnableAppxstart(see Appx.~\ref{appx: parameters}).
\else
(see Appx. A.9).
\EnableAppxend
%------------------------------------------------------------

\noindent{\textbf{Overall algorithm.}} \label{main: algo}
The entire procedure of MACMA is presented as pseudo-code in Algorithm~\ref{sec: algorithm}.

\begin{algorithm}[tb]
\caption{MACMA (Measuring Acoustics with Collaborative Multiple Agents)}
\label{sec: algorithm}
\textbf{Input}: Environment $\mathcal{E}$, \# updates $M$, \# episode $N$, max time steps $T$.\\
\textbf{Parameter}: Stochastic policies $\pi$, initial actor-critic weights $\boldsymbol\theta_0$, initial other weights except for actor-critic weights $\boldsymbol\Omega_0$.\\
\textbf{Output}:Trained weights, $\boldsymbol\theta_{M}$ and $\boldsymbol\Omega_{M}$.

\begin{algorithmic}[1] %[1] enables line numbers
    \FOR {$i$=1, 2, ... $M$}
        \STATE // Run policy $\pi_{\theta_{i-1}}$ for $N$ episodes $T$ time steps \;  \\
        \STATE 	$\{(o_{t,i},\,h_{t-1,i},\,a_{t,i},\,r_{t,i})\} \leftarrow \text{roll}(\mathcal{E}, \pi_{\boldsymbol\theta_{i-1}}, T)$\; \\
        \STATE Compute advantage estimates \; \\
        \STATE RIR prediction and environmental reward assignment\; \\
        \STATE //  Optimize w.r.t. $\boldsymbol\theta$ and $\boldsymbol\Omega$\; \\
        \STATE $\boldsymbol\theta_{i}, \boldsymbol\Omega_i \leftarrow$ new $\boldsymbol\theta$ and $\boldsymbol\Omega$ from PPO algorithm w.r.t. minimizing \Eqref{eq: loss-total} \;  
    \ENDFOR
\end{algorithmic}
\end{algorithm}

% \section{EXPERIMENTS}
\section{Experiments}

We adopt the commonly used 3D environments collected using the SoundSpaces platform~\cite{SoundSpaces} and Habitat simulator~\cite{habitat-sim}.
They are publicly available as several datasets: Replica~\cite{replica}, Matterport3D~\cite{matterport3d} and SoundSpaces (audio)~\cite{SoundSpaces}.
Replica contains 18 environments in the form of grids (with a resolution of 0.5 meters) constructed from accurate scans of apartments, offices, and hotels. 
Matterport3D has 85 scanned grids (1-meter resolution) of indoor environments like personal homes.
To measure the RIR in Replica~\cite{replica} and Matterport3D~\cite{matterport3d}, we let two agents move a certain number of steps (250 and 300 steps for Replica and Matterport3D, respectively) throughout the scene and plan a measuring path.
At every time step, the two agents measure the RIR while moving.
The experimental procedure contains several phases:
a) we pretrain generator $D_{r}$ under the setting $\mathcal{L} = \mathcal{L}^{m}$ ($w_{m}$ = 1.0 and $w_{\xi}$ = 0.0) with random policy for both agent 0 and agent 1 in the training split,
b) we train and validate every baseline with the generator $D_{r}$ fine-tune together in the training and validation split,
c) we test every baseline in the test split.
%---------------------------------------------
%---------------------------------------------
MACMA is benchmarked towards several baselines: {\bf Random}, {\bf Nearest neighbor}, {\bf Occupancy}~\cite{Occupancy} and {\bf Curiosity}~\cite{curiosity}.
{\bf Random} uniformly samples one of three actions and executes \textit{Stop} when it reaches maximum steps.
{\bf Nearest neighbor} predict from closest experience \EnableAppxstart(Appx.~\ref{appx: Nearest-neighbor}).
\else (Appx. A.3). \EnableAppxend
{\bf Occupancy} orient to occupy more area, making the area of $\Gamma_{ABCD}$ in Figure~\ref{fig: convex} larger. 
{\bf Curiosity} strives to visit nodes that have not been visited already in the current episode.
%------------------------------------------------------------------------------
To evaluate different methods, we adopt the evaluation metrics {\bf CR} (coverage rate), {\bf PE} (prediction error),{\bf WCR} (weighted coverage rate),{\bf RTE} (RT60 Error) and {\bf SiSDR} (scale-invariant signal-to-distortion ratio), among which \textbf{WCR} is the most important evaluation metric since it is a trade-off between encouraging high prediction accuracy and more exploration. 
{\bf CR} is the ratio of the number of visited nodes by agents and the number of nodes in the current episode.
$CR=\nicefrac{N_{v}}{N_{e}}$, where $N_{v}$ is the total number of unique nodes that two agents have visited together, and $N_{e}$ is the total number of all individual nodes in the current episode. 
$\text{PE} =\Delta(W_{t},\,\hat{W}_{t})$, where $W_{t}$  (\Eqref{metric: stft-distance}) is the ground truth RIR, $\hat{W}_{t}$ is the predicted RIR.
$\text{WCR} = ( 1.0 - \lambda ) * \text{CR} + \lambda * ( 1.0 - \text{PES})$, where PES stands for Scaled Prediction Error. $\text{PES} = 2/(1+\exp(-\text{PE})) - 1.0$, where $0 \le \lambda \le 1.0$ is a hyper-parameter.
{\bf RTE} describes the difference between the ground truth RT60 value and the predicted one.
SiSDR = $10 \log_{10} \nicefrac{\lVert X_{T} \rVert^{2} }{ \lVert X_{E}  \rVert^{2} }$, where $\lVert X_{E} \rVert^{2}$ is the error vector and $\lVert X_{T} \rVert^{2}$ is the ground truth vector.
\EnableAppxstart
We select the hyperparameters in grid search and the details are in Appx.~\ref{appx:hyperparameterselection}.
\else
We select the hyperparameters in grid search and the details are in Appx. A.4.
\EnableAppxend
%---------------------------------------------
\begin{table*}[t!]
    % \addtolength{\tabcolsep}{-5pt}
    % \renewcommand{\arraystretch}{1.2}
    \centering
    \resizebox{0.90\linewidth}{!}{
    \begin{tabular}{l|ccccc|ccccc}
        \hline
        \multirow{2}{*}{Model} &  \multicolumn{5}{c|}{Replica}  & \multicolumn{5}{c}{Matterport3D}      \\
        \cline{2-11}
          & WCR ($\uparrow$) & PE ($\downarrow$)   & CR ($\uparrow$) & RTE ($\downarrow$)  & SiSDR ($\uparrow$)  & WCR ($\uparrow$) & PE ($\downarrow$)   & CR ($\uparrow$) & RTE ($\downarrow$)  & SiSDR ($\uparrow$)  \\
        \hline
        Random           & 0.3103   &   5.4925 & 0.3439 & 14.7427 & 20.3534   &  0.2036 & 5.5552 & 0.2254 & 23.5281 & 12.3042     \\
        Nearest neighbor & 0.3444   &   5.4533 & 0.3817 & 14.0269 & 22.0135   &  0.2099 & 5.3342 & 0.2321 & 28.8765 & 15.2351   \\
        Occupancy        & 0.4464   &   3.7224 & 0.4907 & 12.5532 & 23.0666   &  0.2225 & 4.5327 & 0.2449 & 20.3399 & 18.3848  \\
        Curiosity        & 0.4327   &   3.4883 & 0.4742 & \textbf{10.9565} & \textbf{23.8669}   & 0.2111 & 4.4255 & 0.2319 & 29.5572 & 20.0031   \\
        MACMA (Ours)    & \textbf{0.6977} &   \textbf{3.2509}  & \textbf{0.7669} & 13.8896 & 23.6501   & \textbf{0.3030} & \textbf{4.0113} & \textbf{0.3327} & \textbf{15.9338} & \textbf{21.3187}  \\
        \hline
    \end{tabular}
    }
    \caption{The results of quantitative comparison between our proposed method (MACMA) and baselines.}
    \label{tab: com-baselines}
\end{table*}
% --------------------------------------------------------------------
%-------------------------------------------------------------------------------
\begin{figure*}[t!]
\centering
\includegraphics[width=0.95\textwidth]{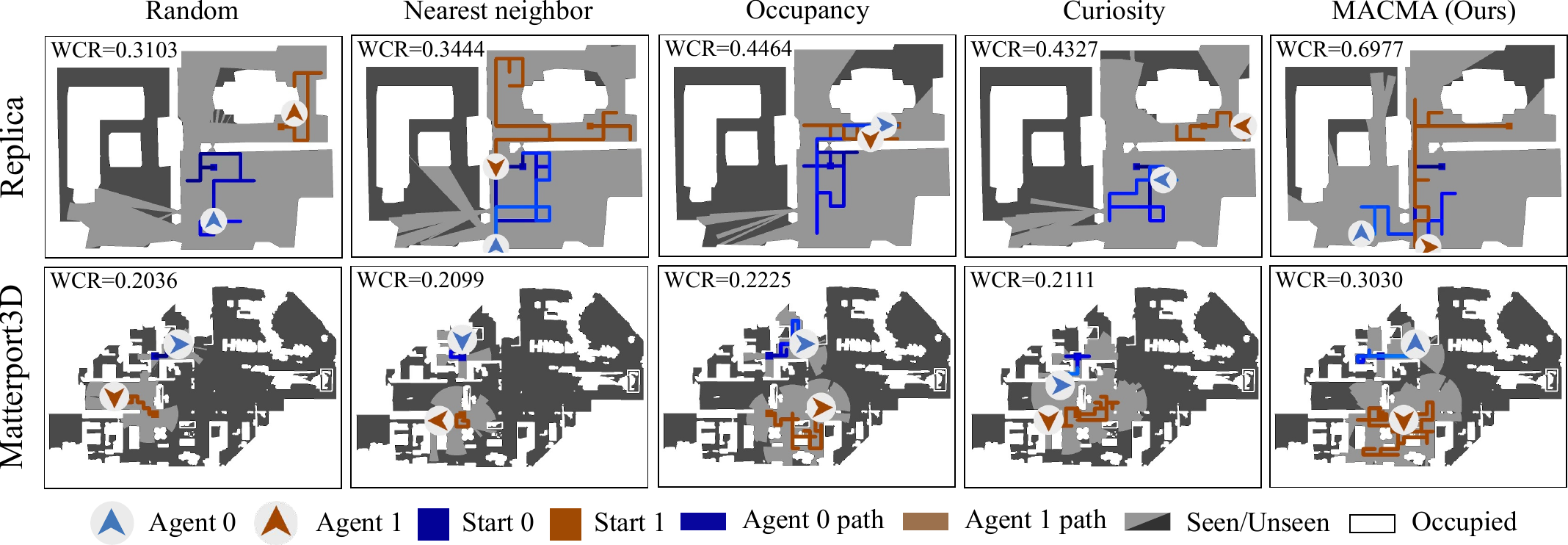}
\caption{\small Visualization of the navigation trajectories by the end of a particular episode from Replica (top row) and Matterport3D (bottom row) dataset. 
Higher WCR values and bigger ``seen'' areas (colored in light-grey) indicate better performances.
}
\label{fig: traj}
\end{figure*}

\subsection{Experimental Results}

Results are averages of 5 tests with different random seeds.

\noindent{\textbf{Quantitative comparison of the two datasets.}}
Results are runs on two datasets under the experimental settings: $\alpha^{\xi}$=1.0, $\alpha^{\zeta}$=1.0, $\alpha^{\psi}$=-1.0, $\alpha^{\phi}$=1.0, $\kappa$=2, $\lambda$=0.1, $\rho$=-1.0.
As seen from Table~\ref{tab: com-baselines}, on the Replica dataset, MACMA achieves the best results on the metrics WCR, PE and CR. 
But the Curiosity model has the best results on the metrics RTE and SiSDR. 
The Curiosity model encourages agents to visit more new nodes near them, which drive up the robots' exploration ability (it improves performance on metrics RTE and SiSDR) while reducing their team's performance (it reduces performance on the metric CR).
The Occupancy model (ranks as the second over CR) motivates the exploratory ability of the entire group (the group of agent 0 and agent 1) but ignores their individual exploration performance (the ranks over the metrics of RTE and SiSDR lower than that of CR).
MACMA combines the group exploration ability and the individual exploration ability, achieving a good trade-off between the two abilities, so that the group exploration ability of MACMA has increased by a large margin (e.g. over the CR metric) and finally won the championship on the WCR metric.
On the Matterport3D dataset, MACMA achieves the best results on all metrics.
As a result, we can conclude that MACMA quantitatively outperforms baselines over both datasets.

\noindent{\textbf{Qualitative comparison on exploration capability.}} 
Figure~\ref{fig: traj} shows the navigation trajectories of agent 0 and agent 1 for different algorithms by the end of a particular episode from the Replica (top row) and Matterport3D (bottom row) dataset.
The light-gray areas in Figure~\ref{fig: traj} indicate the exploration field of the robots. 
We observe that MACMA tends to explore the most extensively compared to the other baselines.
Particularly, there are three rooms in the entire scene in Replica, and MACMA is the only method that managed to traverse all three rooms using the same number of time steps as baselines.
% --------------------------------------------------------------------

\begin{figure*}[tbh!]
\centering
\includegraphics[width=\textwidth]{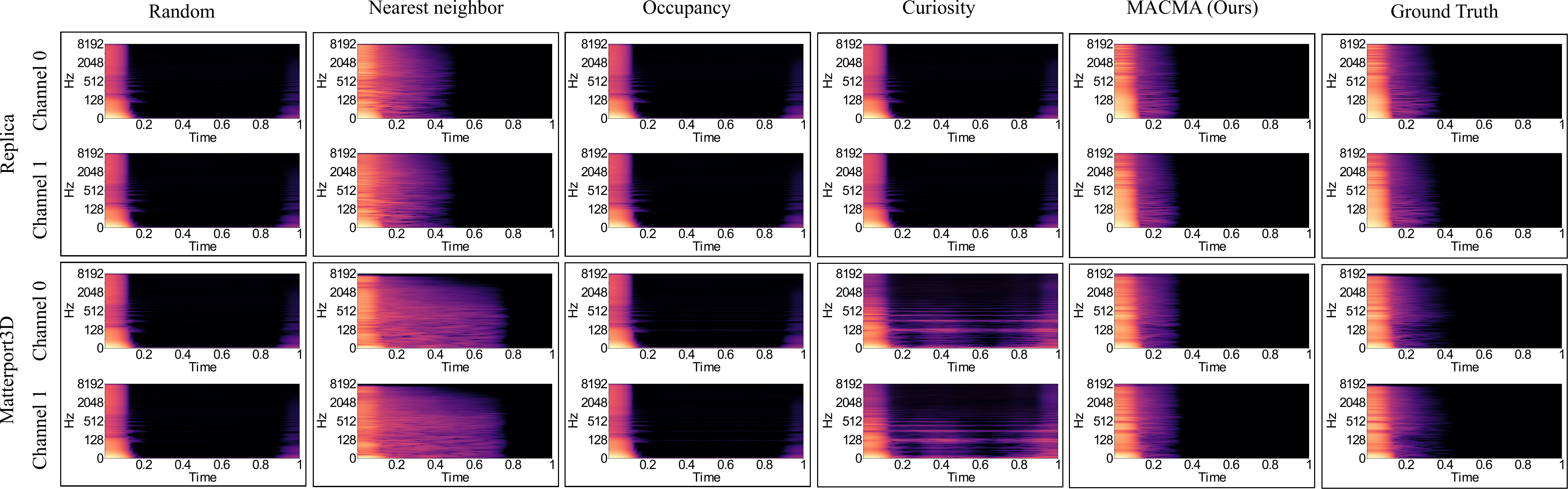}
\caption{\small Qualitative comparison of RIR prediction (Binaural RIR with channel 0 and channel 1) in spectrogram from Replica (top row) and Matterport3D (bottom row) dataset.
Every column is the result of one model except the last one.
The last column is the ground truth of RIR.
}
\label{fig: spec-com}
\vspace{-12pt}
\end{figure*}

\noindent{\textbf{Qualitative comparison on RIR prediction. }} \label{appx: spec-comp}
We show the spectrograms generated by these models and from the ground truth in Figure~\ref{fig: spec-com}.
%-------------------------------------------------------------------------------
%-------------------------------------------------------------------------------
These binaural spectrograms with channel 0 and channel 1 last for one second (the x-axis is the time axis).
The spectrogram of the RIR from both Random's and Occupancy's generation have fewer color blocks than the ground truth between 0.2 seconds and 0.4 seconds and more color blocks than the ground truth between 0.8 seconds and 1 second.
The spectrogram of the RIR from the Nearest neighbor's generation has more colored regions than the ground truth spectrogram.
The spectrogram of the RIR from Curiosity's prediction has fewer color blocks than the ground truth between 0.2 seconds and 0.4 seconds.
At the same time, the spectrogram of the RIR from Curiosity's generation has more color blocks than the ground truth between 0.8 seconds and 1 second in the Replica dataset.
And the spectrogram of the RIR from Curiosity's prediction has more colored regions than the ground truth spectrogram in the Matterport3D dataset.
The spectrogram of the generated RIR from MACMA (Ours) is the closest to the ground truth spectrogram.
In conclusion, from a qualitative human visual point of view, the spectral quality of the RIRs generated by our model is the best.
\EnableAppxstart
Additionally, in Appx.~\ref{appx: rir-comp}, we show that the RIR's  quality in the waveform of the RIRs generated by our model is also superior.
\else
Additionally, in Appx. A.5, we show that the RIR's  quality in the waveform of the RIRs generated by our model is also superior.
\EnableAppxend

\subsection{Ablation Studies}

\noindent{\textbf{Ablation on modality. } }
Results are run on dataset Replica under the experimental settings of
$\alpha^{\xi}$=1.0, $\alpha^{\zeta}$=1.0, $\alpha^{\psi}$=-1.0, $\alpha^{\phi}$=1.0, $\kappa$=2, $\lambda$=0.1, $\rho$=-1.0.
As shown in Table~\ref{tab: abla-modality}, RGBD (vision with RGB images and Depth input) seems to be the best choice.

\noindent{\textbf{More ablations. } }
We explore the relationship between modality importance, action selection, and RIR measurement accuracy in \EnableAppxstart Appx.~\ref{appx: vap-action} \else Appx. A.6 \EnableAppxend and \EnableAppxstart Appx.~\ref{appx: vap-rir}. \else Appx. A.7. \EnableAppxend
%\EnableAppxstart　
We present the extension model MACMARA (MACMA with a dynamic Reward Assignment module) in Appx.~\ref{appx-exp-m2mrirra}. 
More ablation studies on memory size $\kappa$ and the reward component can be found in Appx.~\ref{appx:hyperparameterselection}.
%
%\else
%We present the extension model MACMARA (MACMA with a dynamic Reward Assignment module) in Appx. B.5. 
%More ablation studies on memory size $\kappa$ and the reward component can be found in Appx. A.4.
%\EnableAppxend

% ------------------------------------------
\begin{table}[h!]
    \centering
    \resizebox{0.95\linewidth}{!}{
    \begin{tabular}{l|ccccc}
        \hline
        Vision & WCR ($\uparrow$) &  PE ($\downarrow$)  & CR ($\uparrow$) & RTE ($\downarrow$)  & SiSDR ($\uparrow$)   \\
        \hline
            Blind        & 0.5020 &  3.4966 & 0.5512 & 14.2049 & 23.0903  \\
            RGB          & 0.5930 &  3.8204 & 0.6541 & 15.5897 & \textbf{23.7713} \\
            Depth        & 0.5068 &  3.4927 & 0.5566 & 29.6905 & 23.5089  \\
            RGBD         & \textbf{0.6977} &  \textbf{3.2509} & \textbf{0.7669} & \textbf{13.8896} & 23.6501   \\
        \hline
    \end{tabular}
     }
    \caption{Ablation on modality.}
    \label{tab: abla-modality}
    \vspace{-0.05in}
\end{table}

% \section{CONCLUSION}
\section{Conclusion}

In this work, we propose a novel task where two collaborative agents learn to measure room impulse responses of an environment by moving and emitting/receiving signals in the environment within a given time budget. To tackle this task, we design a collaborative navigation and exploration policy. Our approach outperforms several other baselines on the environment's coverage and prediction error. 
A known limitation is that we only explored the most basic setting, one listener (receiver), and one source (emitter), and did not study the settings with two or more listeners or sources.
Another limitation of our work is that our current assessments are conducted in a virtual environment.
It would be more meaningful to evaluate our method on real-world cases, such as a robot moving in a real house and learning to measure environmental acoustics collaboratively.
Lastly, we have not considered semantic information about the scene in policy learning.
Incorporating semantic information about the scene into policy learning would be more meaningful.
The above three are left for future exploration.

\section*{Ethical Statement} \label{appendix: ethics}
This research follows \href{https://ijcai-23.org/call-for-papers/}{IJCAI's ethics guidelines} and does not involve human subjects or privacy issues with the dataset. The dataset is publicly available.

% \section*{ACKNOWLEDGMENTS}
\section*{Acknowledgments}
We thank the IJCAI 2023 reviewers for their insightful feedback. This work is supported by multiple projects, including the Sino-German Collaborative Research Project {\it Crossmodal Learning} (NSFC62061136001/DFG SFB/TRR169), the National Natural Science Foundation of China (No. 61961039), the National Key R\&D Program of China (No. 2022ZD0115803), the Xinjiang Natural Science Foundation (No. 2022D01C432, No. 2022D01C58, No. 2020D01C026, and No. 2015211C288), and partial funding from the THU-Bosch JCML Center.

%%%%%%%%%%%%%%%%%%%%%%%%%%%%%%%%%%%%%%%%%%%%%%%%%%%%%%%%%%%%%%%%%%%%%%%%
% \clearpage
\EnableMainend
%%%%%%%%%%%%%%%%%%%%%%%%%%%%%%%%%%%%%%%%%%%%%%%%%%%%%%%%%%%%%%%%%%%%%%%%
\EnableAppxstart  % start of appx
%%%%%%%%%%%%%%%%%%%%%%%%%%%%%%%%%%%%%%%%%%%%%%%%%%%%%%%%%%%%%%%%%%%%%%%%
% The formulas start at 12, figures at 7, and tables at 3 in the appendix section.
\setcounter{equation}{11}
\setcounter{figure}{5}
\setcounter{table}{2}
%--------------------------------------------------
\appendix
% \section*{\LARGE Appendix} \label{sec:appendix}
\section*{APPENDIX} \label{sec:appendix}

%--------------------------------------------------
We provide additional details for MACMA and the extension version model MACMARA (Measuring Acoustics with Collaborative Multiple Agents with a reward assignment module).

\ref{appx: M2MRIR}:~~
MACMA

\hspace{0.2in}\ref{appx:theoretical}:~~
The theoretical point of view.

\hspace{0.2in}\ref{appx: structure-edr}:~~
Implementation of $E_{r}$, $E_{m}$ and $D_{r}$ in details.

\hspace{0.2in}\ref{appx: Nearest-neighbor}:~~
Implementation of nearest neighbor in detail. 

\hspace{0.2in}\ref{appx:hyperparameterselection}:~~
Hypter-parameter selection experiments.

\hspace{0.2in}\ref{appx: exp}:~~
More experiments for MACMA.

\hspace{0.2in}\ref{appx: vap-action}:~~
On modality importance for action selection.

\hspace{0.2in}\ref{appx: vap-action}:~~
On modality importance for action selection.

\hspace{0.2in}\ref{appx: state-action}:~~
Visualization of agents' state features.

\hspace{0.2in}\ref{appx: parameters}:~~
Algorithm parameters.

\ref{appx: m2mrirra}:~~
MACMARA

\hspace{0.2in}\ref{appx: overview-m2mrirra}:~~
The overview of MACMARA.

\hspace{0.2in}\ref{appx: env-reward-assign-m2mrirra}:~~
Reward assignment for MACMARA.

\hspace{0.2in}\ref{appx: loss-m2mrirra}:~~
The loss of MACMARA.

\hspace{0.2in}\ref{appx: formalation-m2mrirra}:~~
Formalation of MACMARA.

\hspace{0.2in}\ref{appx-exp-m2mrirra}:~~
Experimental results of MACMARA.

\hspace{0.2in}\ref{appx: m2mrirra-parameters}:~~
Algorithm parameters in MACMARA.

%--------------------------------------------------
%$$$$$$$$$$$$$$$$$$$$$$$$$$$$$$$$$$$$$$$$$$$$$$$$$$$$$$$$$$$$$$$$$$$$$$$$$$$$$
% \clearpage
%%%%%%%%%%%%%%%%%%%%%%%%%%%%%%%%%%%%%%%%%%%%%%%%%%%%%%%%%%%%%%%%%%%%%%%%
\section{MACMA} \label{appx: M2MRIR}

\subsection{The Theoretical Point of View}\label{appx:theoretical}

Below we provide some theoretical points of view for formulation.

\noindent{\textbf{Monotonic value function factorization theory. }} Monotonic value function factorization for deep multi-agent reinforcement learning is proposed in QMIX~\cite{maqmix}. 
This theorem is cited as follows.
\begin{lemma} \label{lemma1}
If $\forall i \in \mathbf{N} \equiv\{1,2, \ldots, n\}, \frac{\partial Q_{tot}}{\partial Q_{i}} \geq 0$ then
\begin{equation} \label{eq: lemma1-appx}
\resizebox{.65\hsize}{!}{
$
    \mathop{\arg\max}\limits_{\mathbf{u}} Q_{to t}(\boldsymbol{\tau}, \mathbf{u})=\left(\begin{array}{c}
    \mathop{\arg\max}\limits_{\mathbf{u^{1}}} Q_{1}\left(\tau^{1}, u^{1}\right) \\
    \vdots \\
    \mathop{\arg\max}\limits_{\mathbf{u^{n}}} Q_{n}\left(\tau^{n}, u^{n}\right)
    \end{array}\right)$}\,,
\end{equation}
\end{lemma}
\noindent
where $Q_{tot}(\boldsymbol{\tau}, \mathbf{u})$ is a joint action-value function, $\boldsymbol{\tau}$ is a history of joint action, and $\mathbf{u}$ is a joint action.

\noindent
The proof of this theorem is provided as:
\begin{proof}
Since $\frac{\partial Q_{tot}}{\partial Q_{i}} \geq 0$ for $\forall i \in \mathbf{N}$, the following holds for any $\left(u^{1}, \ldots, u^{n}\right)$ and the mixing network function $Q_{tot}(\cdot)$ with $n$ arguments:
\begin{equation}
\resizebox{0.95\hsize}{!}{$
    \begin{array}{l}
    Q_{tot}\left(Q_{1}\left(\tau^{1}, u^{1}\right), \ldots, Q_{i}\left(\tau^{i}, u^{i}\right), \ldots, Q_{n}\left(\tau^{n}, u^{n}\right)\right) \\
    \leq Q_{tot}\left(\mathop{\max}\limits_{u^{1}} Q_{1}\left(\tau^{1}, u^{1}\right), \ldots, Q_{i}\left(\tau^{i}, u^{i}\right), \ldots, Q_{n}\left(\tau^{n}, u^{n}\right)\right)  \quad \quad \ldots  \\
    \leq Q_{tot}\left(\mathop{\max}\limits_{u^{1}} Q_{1}\left(\tau^{1}, u^{1}\right) ,\,\ldots,\, \mathop{\max}\limits_{u^{i}} Q_{i}\left(\tau^{i}, u^{i}\right) \ldots,\, Q_{n}\left(\tau^{n}, u^{n}\right)\right)  \quad \quad \ldots  \quad \quad \\ 
    \leq Q_{tot}\left(\mathop{\max}\limits_{u^{1}} Q_{1}\left(\tau^{1}, u^{1}\right), \ldots, \mathop{\max}\limits_{u^{i}} Q_{i}\left(\tau^{i}, u^{i}\right),\,\ldots,\, \mathop{\max}\limits_{u^{n}} Q_{n}\left(\tau^{n}, u^{n}\right)\right).
    \end{array}
$}
\end{equation}
\noindent
Therefore, the maximum of the mixing network function is:\\ 
\resizebox{0.75\hsize}{!}{
$
\left(\mathop{\max}\limits_{u^{1}} Q_{1}\left(\tau^{1}, u^{1}\right), \ldots, \mathop{\max}\limits_{u^{i}} Q_{i}\left(\tau^{i}, u^{i}\right),\,\ldots,\, \mathop{\max}\limits_{u^{n}} Q_{n}\left(\tau^{n}, u^{n}\right)\right)
$}. Thus, 
\begin{equation}
\resizebox{.85\hsize}{!}{$
    \begin{aligned}
        \mathop{\max}\limits_{\mathbf{u}} Q_{tot}(\boldsymbol{\tau},\,\mathbf{u}):&=\max _{\mathbf{u}=\left(u^{1},\,\ldots, u^{n}\right)} Q_{tot}\left(Q_{1}\left(\tau^{1}, u^{1}\right), \ldots, Q_{n}\left(\tau^{n}, u^{n}\right)\right) \\
        &=Q_{tot}\left(\max_{u^{1}} Q_{1}\left(\tau^{1}, u^{1}\right), \ldots, \max_{u^{n}} Q_{n}\left(\tau^{n}, u^{n}\right)\right).
    \end{aligned}
$}
\end{equation}
Letting,
\begin{equation}
\resizebox{.55\hsize}{!}{
$
    \mathbf{u}_{\star}=\left(u_{\star}^{1},\,\ldots,\, u_{\star}^{n}\right)=\left(\begin{array}{c}
    \mathop{\arg\max}\limits_{u^{1}} Q_{1}\left(\tau^{1}, u^{1}\right) \\
    \vdots \\
    \mathop{\arg\max}\limits_{u^{n}} Q_{n}\left(\tau^{n}, u^{n}\right)
    \end{array}\right),
$}
\end{equation}
\noindent
we have that
\begin{equation}
\resizebox{.85\hsize}{!}{$
    \begin{aligned}
        Q_{tot}\left(Q_{1}\left(\tau^{1}, u_{\star}^{1}\right),\,\ldots,\,Q_{n}\left(\tau^{n}, u_{\star}^{n}\right)\right)
        &=Q_{tot}\left(\mathop{\max}\limits_{u_{1}} Q_{1}\left(\tau^{1}, u^{1}\right),\,\ldots,\, \mathop{\max}\limits_{u_{n}} Q_{n}\left(\tau^{n}, u^{n}\right)\right) \\
        &=\mathop{\max}\limits_{\mathbf{u}} Q_{tot}(\boldsymbol{\tau},\,\mathbf{u}).
    \end{aligned}
$}
\end{equation}
\noindent
Hence,
$
\mathbf{u}_{\star}=\mathop{\arg\max}\limits_{\mathbf{u}} Q_{tot}(\boldsymbol{\tau}, \mathbf{u})
$, which proves \Eqref{eq: lemma1-appx}. 
\end{proof}

\noindent{\textbf{The property of the decomposition theorem of the monotone value function. }} According to Lemma~\ref{lemma1}, we can get the following property: 
\EnableMainstart
\begin{property} \label{prop1}
If $\forall i \in \{{\omega},\,{\nu}\}, \frac{\partial G(\pi^{\omega},\,\pi^{\nu},\,r)}{\partial G(\pi^{i},\,r)} \geq 0$ then \Eqref{eq: total-ques} has an optimal solution.
\end{property}
\else
\begin{property} \label{prop1}
If $\forall i \in \{{\omega},\,{\nu}\}, \frac{\partial G(\pi^{\omega},\,\pi^{\nu},\,r)}{\partial G(\pi^{i},\,r)} \geq 0$ then Equation 1 has an optimal solution.
\end{property}
\EnableMainend

\noindent{\textbf{Construction of objective function of proposed problem. }} 
\EnableMainstart
Inspired by \textit{Value Decomposition Networks} (VDNs)~\cite{mavdn} and QMIX~\cite{maqmix}, we construct the objective function $G(\pi^{\omega},\pi^{\nu},r)$ in \Eqref{eq: total-ques} by combining the non-negative partial reciprocal constraint respond to $G(\pi^{\omega},r)$ and $G(\pi^{\nu},r)$ according Property~\ref{prop2}.
\else
Inspired by \textit{Value Decomposition Networks} (VDNs)~\cite{mavdn} and QMIX~\cite{maqmix}, we construct the objective function $G(\pi^{\omega},\pi^{\nu},r)$ in Equation 1 by combining the non-negative partial reciprocal constraint respond to $G(\pi^{\omega},r)$ and $G(\pi^{\nu},r)$ according Property~\ref{prop2}.
\EnableMainend

\EnableMainstart
\begin{property} \label{prop2}
Since $w^{\omega}=\frac{\partial G(\pi^{\omega},\,\pi^{\nu},\,r)}{\partial G(\pi^{\omega},\,r)} > 0$ and $w^{\nu}=\frac{\partial G(\pi^{\omega},\,\pi^{\nu},\,r)}{\partial G(\pi^{\nu},\,r)} > 0$ satisfy the conditions of Property~\ref{prop1}, so \Eqref{eq: total-ques} has an optimal solution.
\end{property}
\else
\begin{property} \label{prop2}
Since $w^{\omega}=\frac{\partial G(\pi^{\omega},\,\pi^{\nu},\,r)}{\partial G(\pi^{\omega},\,r)} > 0$ and $w^{\nu}=\frac{\partial G(\pi^{\omega},\,\pi^{\nu},\,r)}{\partial G(\pi^{\nu},\,r)} > 0$ satisfy the conditions of Property~\ref{prop1}, so Equation 1 has an optimal solution.
\end{property}
\EnableMainend

\subsection{Implementations of $E_{r}$, $E_{m}$ and $D_{r}$ in Details} \label{appx: structure-edr}

Here we provide the structure of $E_{r}$, $E_{m}$ and $D_{r}$ in details.

\noindent{\textbf{The structure of $E_{r}$. }}
\begin{figure}[h]
    % \vspace{-20pt}
      \begin{center}
        \includegraphics[width=0.2\textwidth]{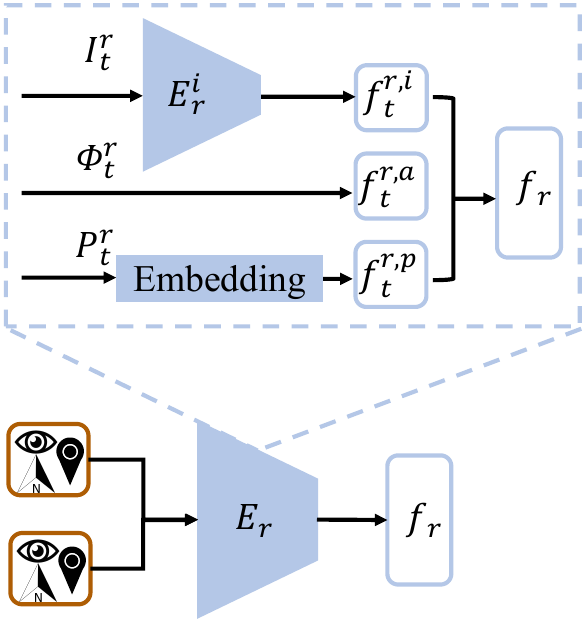}
      \end{center}
      \vspace{-8pt}
      \caption{The structure of $E_{r}$.}
    \label{fig: er-struct}
\end{figure}
As shown in Figure~\ref{fig: er-struct}, at the time step $t$, the visual ($I^{r}_{t}$) part is encoded into visual feature vector $f^{r,\,i}_{t}$ using a CNN encoder.
Visual CNN encoders $E^{i}_{r}$ is constructed in this way (from the input to output layer): \verb|Conv8x8|, \verb|Conv4x4|, \verb|Conv3x3| and a 256-dim linear layer; 
ReLU activations are added between any two neighboring layers.
$P^{r}_{t}$ is embedded by a embedding layer and encoded into feature vector $f_{t}^{r,\,p}$.
Then, we concatenate the three vectors $f_{t}^{r,\,i}$, $f_{t}^{r,\,a}$ and $f_{t}^{r,\,p}$together to obtain the embedding vector $f_{r}$.

\noindent{\textbf{The structure of $E_{m}$. }}
\begin{figure}[h]
    % \vspace{-20pt}
      \begin{center}
        \includegraphics[width=0.2\textwidth]{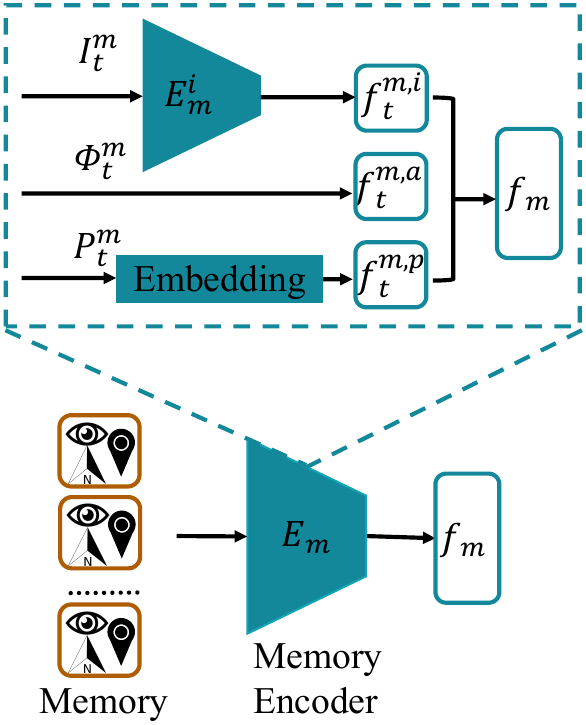}
      \end{center}
      \vspace{-8pt}
      \caption{The structure of $E_{m}$.}
    \label{fig: em-struct}
\end{figure}
As shown in Figure~\ref{fig: em-struct}, at the time step $t$, the visual ($I^{m}_{t}$) part is encoded into visual feature vector $f^{m,\,i}_{t}$ using a CNN encoder.
Visual CNN encoders $E^{i}_{m}$ is constructed in this way (from the input to output layer): \verb|Conv8x8|, \verb|Conv4x4|, \verb|Conv3x3| and a 256-dim linear layer; 
ReLU activations are added between any two neighboring layers.
$P^{m}_{t}$ is embedded by a embedding layer and encoded into feature vector $f_{t}^{m,\,p}$.
Then, we concatenate the three vectors $f_{t}^{m,\,i}$, $f_{t}^{m,\,a}$ and $f_{t}^{m,\,p}$together to obtain the embedding vector $f_{m}$.

\begin{figure}[h]
    % \vspace{-20pt}
      \begin{center}
        \includegraphics[width=0.46\textwidth]{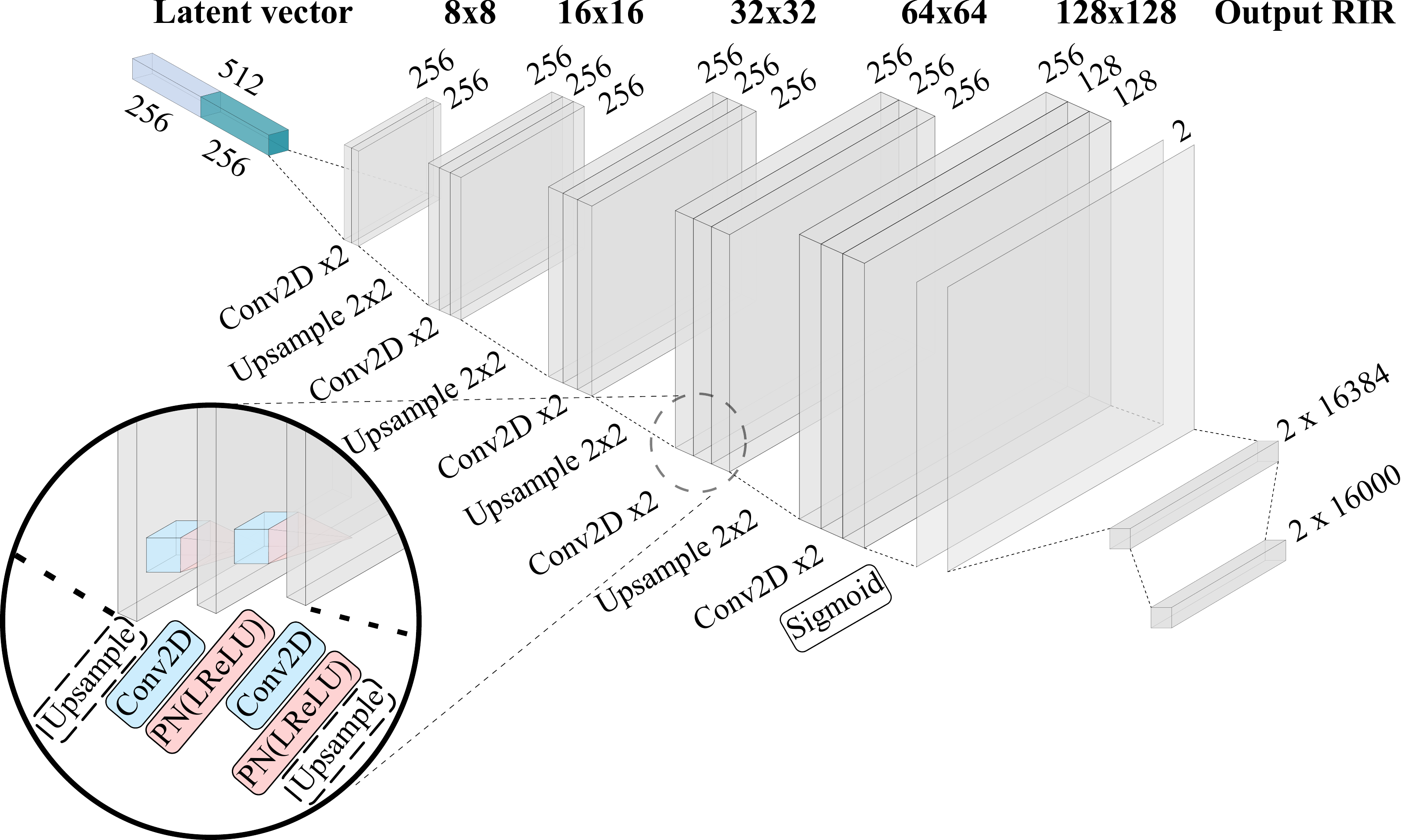}
      \end{center}
      \vspace{-8pt}
      \caption{ Detailed overview of architecture of generator $D_{r}$. }
    \label{fig: dr-generator}
\end{figure}

\noindent{\textbf{The structure of $D_{r}$. }}
As shown in Figure~\ref{fig: dr-generator}, the input of the $D_{r}$ is a 512d latent vector. $D_{r}$ contains several upsampling convolution blocks. 
A leaky rectified linear unit (LReLU), is used after each convolutional layer in $D_{r}$ with the final layer of the generator using a sigmoid activation. 
PN denotes pixel wise normalization, which we use in the generator. The composition of blocks is based on ProGAN~\cite{ProganTK}. 
The final step in $D_{r}$ is a reshape and extractation to make the output with a shape $2 \times 16000$.

\subsection{Implementation Details of Nearest Neighbor} \label{appx: Nearest-neighbor}

When We finished pretrain the generator $D_{r}$, we fixed our model and generate all the latent vector in the train and validation split for a given dataset.
And we store all the latent vector in the form $\langle$scene, scene$\_$id, $f^{i}_{train}$, listener's azimuth, listener's position index, source position index$\rangle$ ( the $i$-th latent vector in the train split).
Then we fixed our pretained model and run in test split for a given dataset, the latent vector $f^{j}_{test}$ generate by Encoder $E_{r}$ and $E_{m}$ (the $j$-th latent vector in the test split).
We calculate similarity $s_{j,\,i}$ by KL (Kullback–Leibler) divergence  between $f^{j}_{test}$ and $f^{i}_{train}$,
$s_{j,\,i} = - D_{KL}(f^{j}_{test} || f^{i}_{train})$.
We get the nearest neighbor by:
\begin{equation} \label{eq: nearest-neighbor}
    i^{\star} = \mathop{\arg\max}\limits_{\mathbf{i \in train }} s_{j,\,i}\,.
\end{equation}
Then we get the record with the nearest neighbor $\langle$scene, scene$\_$id, $f^{i^{\star}}_{train}$, listener's azimuth, listener's position index, source position index$\rangle$.
According to this record, the nearest neighbor model gets the RIR waveform by querying from the train split in a given dataset.

\subsection{Hyper-Parameter Selection Experiments for MACMA}\label{appx:hyperparameterselection}

\noindent{\textbf{The choice of parameter $w^{mse}$. }}
\EnableMainstart
$w^{mse}$ denotes the weight of MSE relative to STFT loss (See in \Eqref{eq: loss-rir-prediction}).
\else
$w^{mse}$ denotes the weight of MSE relative to STFT loss (See in Equation 4).
\EnableMainend
Results (STDEV$\leq$0.01) are averaged over 5 test runs on dataset Replica during pre-training.
Table~\ref{tab: select-weight-mse} shows that larger $w^{mse}$ is helpful for WCR and SiSDR, but it is not good for PE and RTE, so $w^{mse}$ = 1.0 is chosen after a trade-off.
%----------------------------------------------
\begin{table}[!ht]
    \centering
    % \resizebox{0.5\linewidth}{!}{
    \begin{tabular}{c|cccc}
    \hline
        $w^{mse}$       & WCR ($\uparrow$)   & PE ($\downarrow$)    & RTE ($\downarrow$)    & SiSDR ($\uparrow$)     \\
        \hline
        0.0    & 0.1167 & \textbf{2.9037}  & 13.6745 & -21.7048  \\
        0.5    & 0.1602  & 3.1618 & \textbf{13.6742} & -15.9758  \\
        1.0    & \textbf{0.3037} & 3.1188  & 13.8045 & \textbf{24.2508}   \\
    \hline
    \end{tabular}
    % }
    \caption{The choice of parameter $w^{mse}$.}
    \label{tab: select-weight-mse}
\end{table}
%----------------------------------------------

\noindent{\textbf{The choice of parameter $\lambda$. }}
% ------------------------------------------
$\lambda$ is a parameter for metric WCR.
Results (STDEV$\leq$0.01) are averaged over 5 test runs on dataset Replica under the following experimental settings: the model is Curiosity, $\alpha^{\xi}$=1.0, $\alpha^{\zeta}$=1.0, $\alpha^{\psi}$=0.0, $\alpha^{\phi}$=0.0, $\kappa$=2, $\rho$ =-1.0.

It can be seen from Table~\ref{tab: select-lamda} that $\lambda$=0.1 is a good choice.
% ------------------------------------------
% ------------------------------------------
\begin{table}[!ht]
    \centering
    \resizebox{0.95\linewidth}{!}{
    \begin{tabular}{c|ccccc}
    \hline
        $\lambda$ & WCR ($\uparrow$)  & PE ($\downarrow$)  & CR ($\uparrow$) & RTE ($\downarrow$)  & SiSDR ($\uparrow$)   \\
        \hline
        0.1 & \textbf{0.4327} & \textbf{3.4883} & \textbf{0.4742} & \textbf{10.9565} & 23.8669  \\
        0.5 & 0.2364          & 3.7721 & 0.4278 & 10.9687 & 22.9011  \\
        0.9 & 0.2653          & 3.5234 & 0.4732 & 15.9320 & \textbf{24.1206}  \\
    \hline
    \end{tabular}
    }
    \caption{The choice of parameter $\lambda$.}
    \label{tab: select-lamda}
\end{table}
% ------------------------------------------

\noindent{\textbf{The choice of parameter $\alpha^{\psi}$. }} \EnableMainstart
$\alpha^{\psi}$ is a parameter for environmental reward (See in \Eqref{eq: r-cl}).
\else
$\alpha^{\psi}$ is a parameter for environmental reward (See in Equation 11).
\EnableMainend
Results (STDEV$\leq$0.01) are averaged over 5 test runs on dataset Replica under the following experimental settings: the model is Curiosity, $\alpha^{\xi}$=1.0, $\alpha^{\zeta}$=1.0, $\alpha^{\phi}$=0.0, $\kappa$=2, $\rho$=-1.0, $\lambda$=0.1.
It can be seen from Table~\ref{tab: NegConvexLen} that $\alpha^{\psi}$=-1.0 is a good choice.
% --------------------------------------------------------------------
\begin{table}[!ht]
    \centering
    \resizebox{0.95\linewidth}{!}{
    \begin{tabular}{c|ccccc}
    \hline
    $\alpha^{\psi}$ & WCR ($\uparrow$)  & PE ($\downarrow$)  & CR ($\uparrow$) & RTE ($\downarrow$)  & SiSDR ($\uparrow$)   \\
    \hline
        1.0   & 0.4906           & \textbf{3.4624} & 0.5384           & \textbf{12.7661} & \textbf{24.0065}  \\
        -1.0  & \textbf{0.5650}  & 3.4742          & \textbf{0.6211}  & 15.0615          & 23.7385  \\
    \hline
    \end{tabular}
    }
    \caption{The choice of parameter $\alpha^{\psi}$.}
    \label{tab: NegConvexLen}
\end{table}
% --------------------------------------------------------------------

\noindent{\textbf{Ablation on using $\alpha^{\zeta}$=1.0, $\alpha^{\psi}$=-1.0, $\alpha^{\phi}$=1.0. } }
This is ablation study with different reward setting.
\EnableMainstart
We consider whether to use these reward components by using the coefficients $\alpha^{\zeta}$=1.0, $\alpha^{\psi}$=-1.0, $\alpha^{\phi}$=1.0 respectively (See \Eqref{eq: r-cr}).
\else
We consider whether to use these reward components by using the coefficients $\alpha^{\zeta}$=1.0, $\alpha^{\psi}$=-1.0, $\alpha^{\phi}$=1.0 respectively (See Equation 10).
\EnableMainend
Let's take $\alpha^{\psi}$=-1.0 as an example, if the choice is yes, then $\alpha^{\psi}$=-1.0; if the choice is no, then $\alpha^{\psi }$=0.0.
Results (STDEV$\leq$0.01) are averaged over 5 test runs on dataset Replica under the following experimental settings: $\alpha^{\xi}$=1.0, $\kappa$=2, $\rho$=-1.0, $\lambda$=0.1.
It can be seen from the Table~\ref{tab: Curiosity-NegLen-Area} that the best model is: $\alpha^{\zeta}$=1.0, $\alpha^{\psi}$=-1.0, $\alpha^{\phi}$=1.0.
% --------------------------------------------------------------------
\begin{table}[!ht]
    \centering
    \resizebox{0.98\linewidth}{!}{
    \begin{tabular}{ccc|ccccc}
        \hline
        $\alpha^{\zeta}$=1.0 & $\alpha^{\psi}$=-1.0 & $\alpha^{\phi}$=1.0 & WCR ($\uparrow$) & PE ($\downarrow$)  & CR ($\uparrow$) & RTE ($\downarrow$)  & SiSDR ($\uparrow$)   \\
        \hline
        \Checkmark   & \XSolidBrush & \XSolidBrush & 0.4327 & 3.4883 & 0.4742 & \textbf{10.9565} & 23.8669  \\
        \XSolidBrush & \Checkmark   & \XSolidBrush & 0.5215 & 3.7912 & 0.5745 & 14.6120 & 23.1906  \\
        \XSolidBrush & \XSolidBrush & \Checkmark   & 0.4464 & 3.7224 & 0.4907 & 12.5532 & 23.0666  \\
        \Checkmark   & \Checkmark   & \XSolidBrush & 0.5650 & 3.4742 & 0.6211 & 15.0615 & 23.7385  \\
        \Checkmark   & \XSolidBrush & \Checkmark   & 0.4663 & 3.2902 & 0.5101 & 14.7773 & \textbf{23.9743}  \\
        \XSolidBrush & \Checkmark   & \Checkmark   & 0.5468 & 3.2901 & 0.5996 & 13.8064 & 23.8852  \\
        \Checkmark   & \Checkmark   & \Checkmark   &   \textbf{0.6977} & \textbf{3.2509} &   \textbf{0.7669} &    13.8896 &       23.6501   \\
        \hline
    \end{tabular}
    }
    \caption{Ablation on using $\alpha^{\zeta}$=1.0, $\alpha^{\psi}$=-1.0, $\alpha^{\phi}$=1.0.}
    \label{tab: Curiosity-NegLen-Area}
\end{table}
% ------------------------------------------

\noindent{\textbf{The ablations about memory size $\kappa$. } }
$\kappa$ denotes the memory size for the memory
encoder used in our model.
Results (STDEV$\leq$0.01) are averaged over 5 test runs on dataset Replica under the experimental settings:
$\alpha^{\xi}$=1.0, $\alpha^{\zeta}$=1.0, $\alpha^{\psi}$=-1.0, $\alpha^{\phi}$=1.0, $\rho$=-1.0, $\lambda$=0.1.
Table~\ref{tab: number-of-memory} shows that $\kappa$=2 is the best choice.
% --------------------------------------------------------------------
\begin{table}[!ht]
    \centering
    % \resizebox{0.95\linewidth}{!}{
    \begin{tabular}{c|ccccc}
        \hline
        $\kappa $ & WCR ($\uparrow$) & PE ($\downarrow$)  & CR ($\uparrow$) & RTE ($\downarrow$)  & SiSDR ($\uparrow$)   \\
        \hline
             0   & 0.6006 &  \textbf{3.1518} & 0.6582 & 15.3408 & 19.1678  \\
             2   & \textbf{0.6977} &   3.2509 & \textbf{0.7669} & \textbf{13.8896} & 23.6501  \\
             4   & 0.4930 &   3.1751 & 0.5389 & 15.0578 & 22.8031  \\
             8   & 0.5006 &   3.7761 & 0.5513 & 22.6044 & \textbf{24.9619}  \\
            16   & 0.5319 &   3.1821 & 0.5822 & 14.1616 & 22.8854   \\
        \hline
    \end{tabular}
    % }
    \caption{Ablation on $\kappa$.}
    \label{tab: number-of-memory}
\end{table}
% ------------------------------------------

\subsection{More Experiments for MACMA} \label{appx: exp}
We present some results for MACMA below.

\begin{figure*}[t!]
    \centering
    \vspace{0.1in}
    \includegraphics[width=\textwidth]{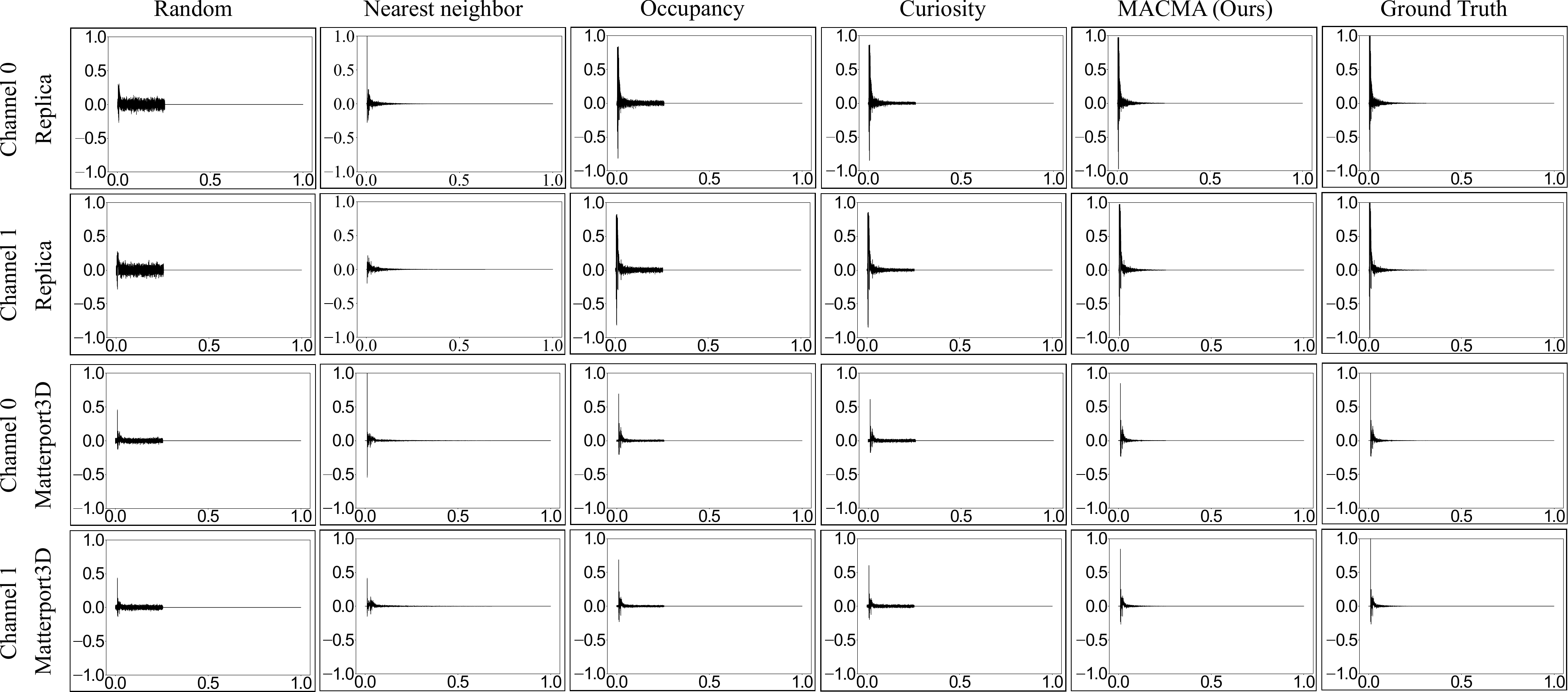}
    \caption{\small Qualitative comparison of RIR prediction (Binaural RIR with channel 0 and channel 1) in waveform from Replica (upper two rows) and Matterport3D (lower two rows).
    The last column is the RIR ground truth.
    }
    \label{fig: rir-com}
    \vspace{-3pt}
\end{figure*}

\noindent{\textbf{Qualitative comparison of RIR prediction in the waveform. } }\label{appx: rir-comp}
In Figure~\ref{fig: rir-com}, we present the RIRs generated by these models together with the ground truth.
These binaural RIRs with channels 0 and 1 last for one second (the x-axis is the time axis).
The shape of the RIRs from Random's prediction is close to ground truth, but the details need to be more accurate both in Replica and Matterport3D datasets.
The RIRs from the Nearest neighbor's prediction for channel 0 are very close to ground truth, but the ones for channel one could be better both in Replica and Matterport3D datasets.
Both channels of the predicted RIR from Occupancy, Curiosity and MACMA are very close to the ground truth. 
Still, the tail part of the predicted RIR from Occupancy and Curiosity is worse than those of MACMA in Replica and Matterport3D datasets.
All in all, from a pure visual point of view, the RIR waveforms generated by our model are of the highest quality.

\begin{figure*}[t!]
\vspace{0.15in}
    \centering
    \begin{subfigure}[h]{.23\textwidth}
      \centering
      \includegraphics[width=0.95\textwidth]{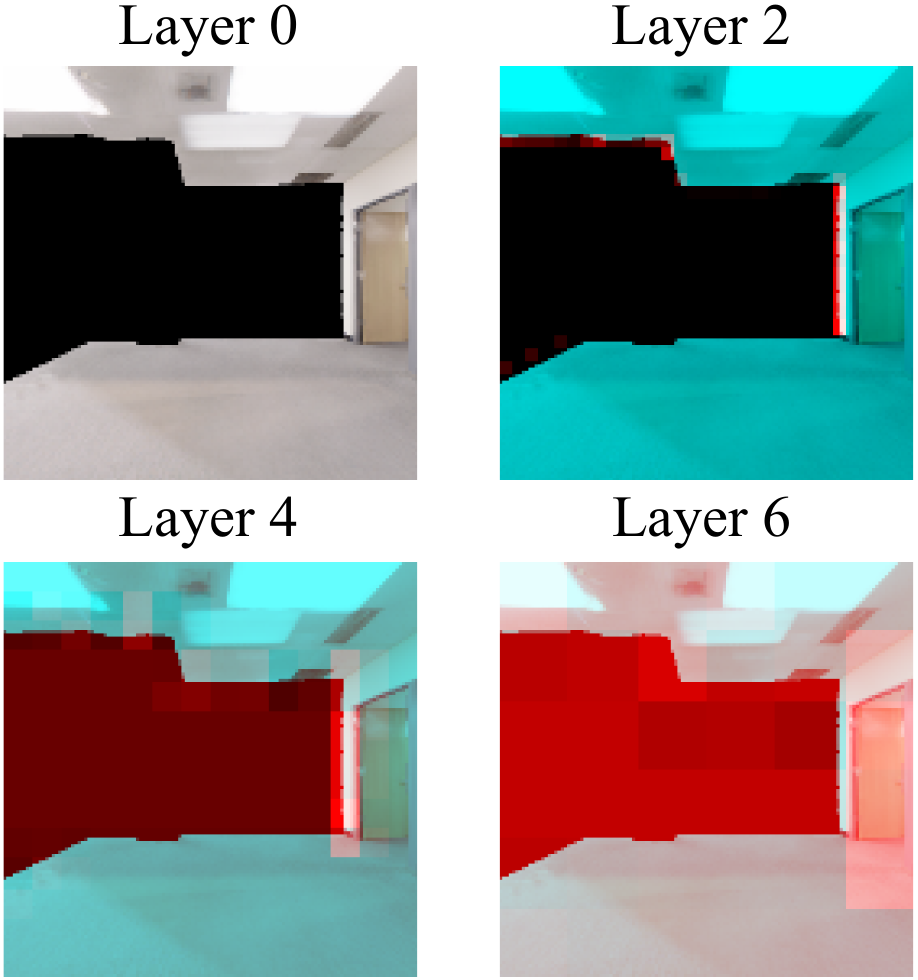}
      \caption{\footnotesize $E_{\omega}$.}
      \label{fig: encoder-omega}
    \end{subfigure}
    \begin{subfigure}[h]{.23\textwidth}
      \centering
      \includegraphics[width=0.95\textwidth]{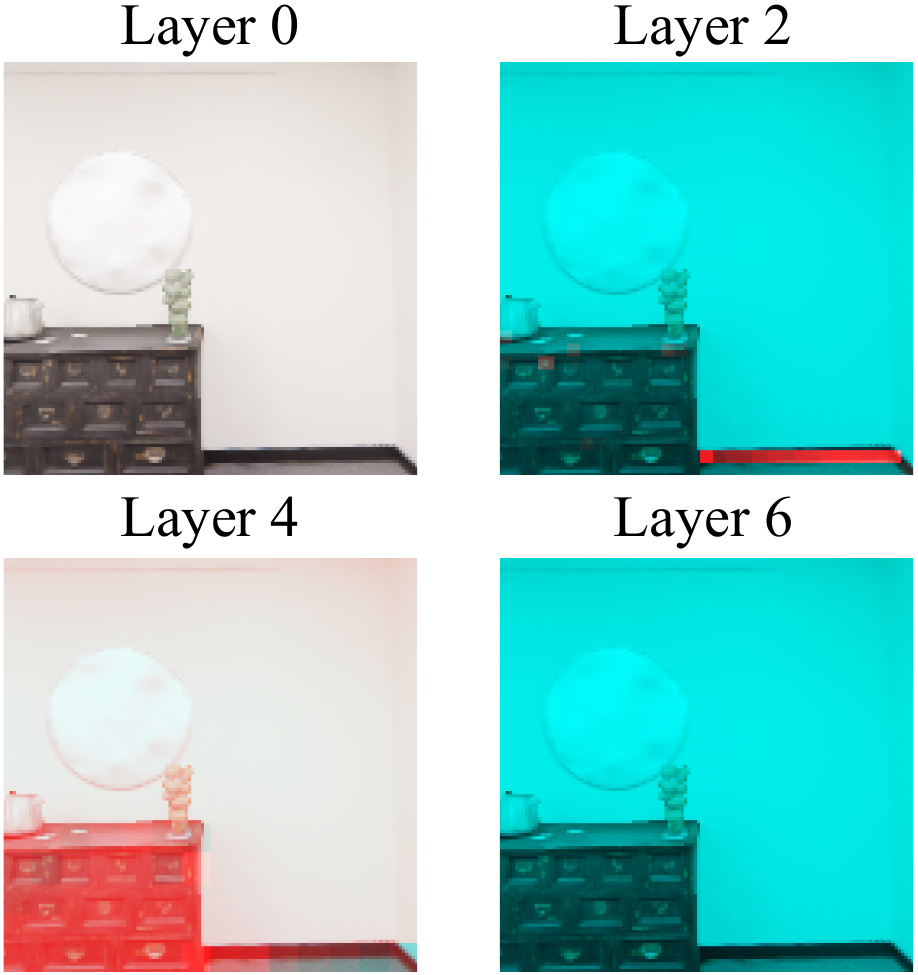}
      \caption{\footnotesize $E_{\nu}$.}
      \label{fig: encoder-nu}
    \end{subfigure}
    \begin{subfigure}[h]{.23\textwidth}
      \centering
      \includegraphics[width=0.95\textwidth]{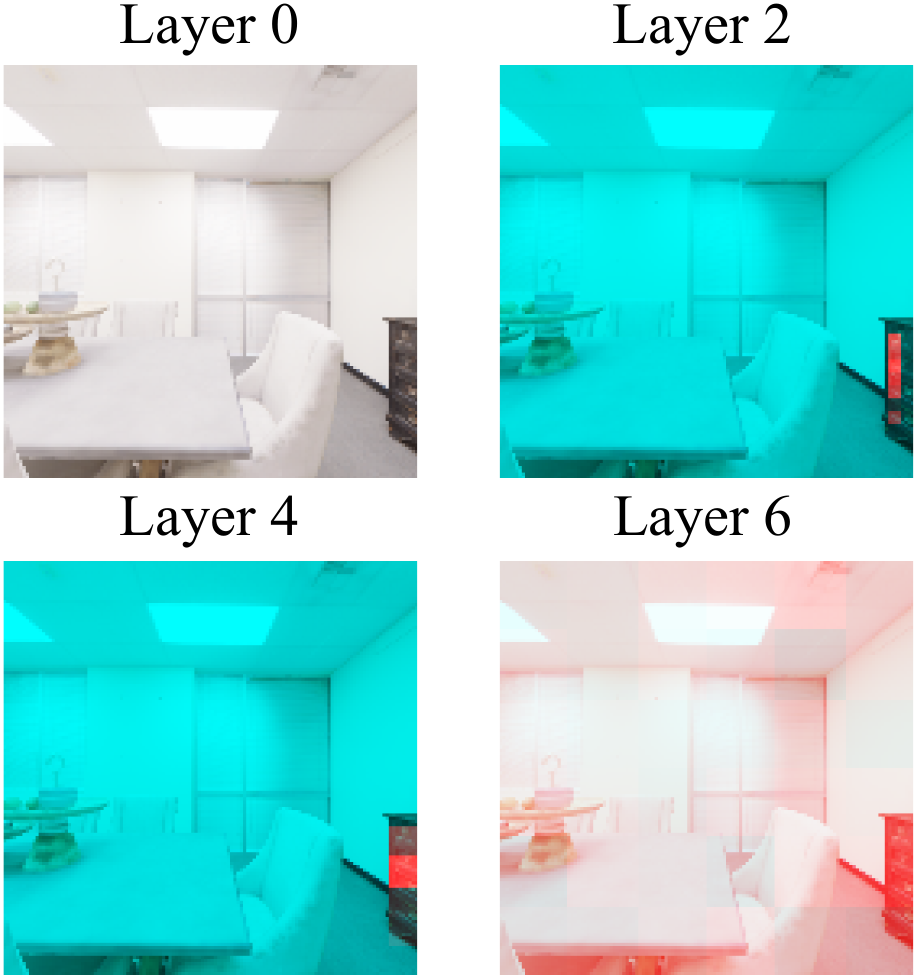}
      \caption{\footnotesize $E_{m}$.}
      \label{fig: encoder-m}
    \end{subfigure}
    \begin{subfigure}[h]{.23\textwidth}
      \centering
      \includegraphics[width=0.95\textwidth]{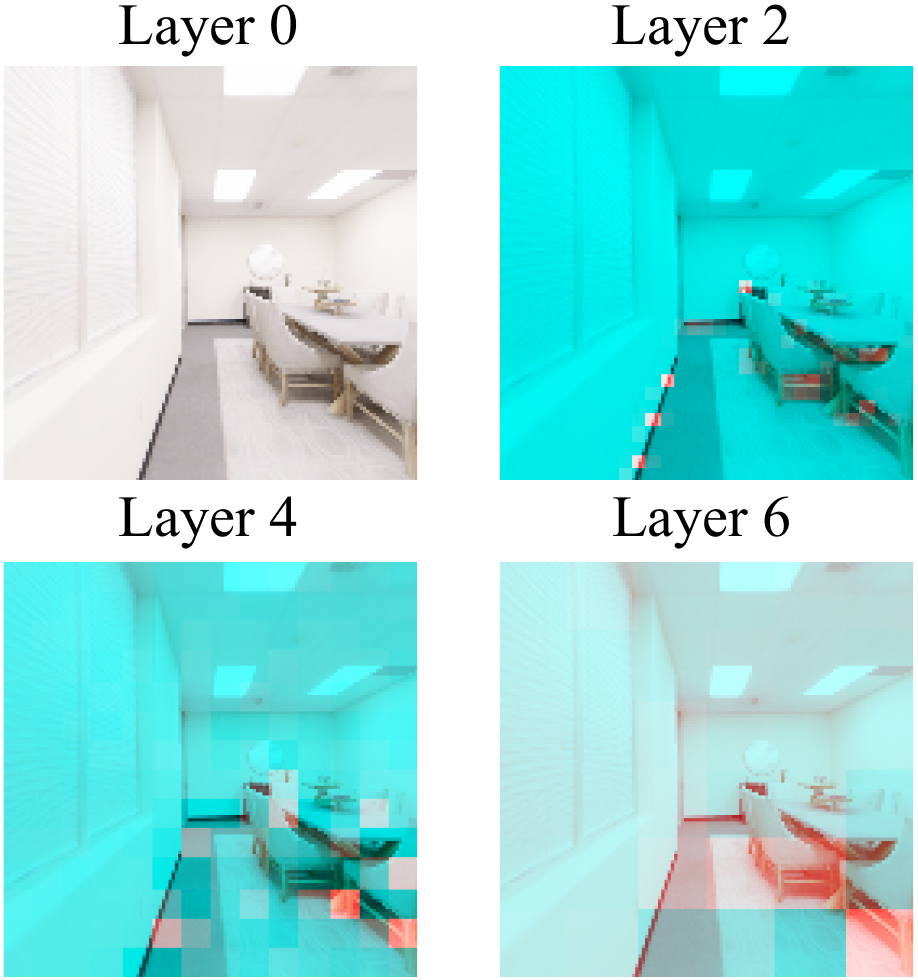}
      \caption{\footnotesize $E_{r}$.}
      \label{fig: encoder-r}
    \end{subfigure}
    \caption{Visualization of the learned features for different encoders (a) $E_{\omega}$,  (b) $E_{\nu}$,  (c) $E_{m}$, (d) $E_{r}$.
    }
    \label{fig: visualization-learned-encoders}
\end{figure*}

\noindent{\textbf{Visualization of learned features and states. }} \label{appx: visual-encoder-decoders}
\EnableMainstart
In MACMA framework (cf. section~\ref{sec: approach}), the encoder $E_{\omega}$, $E_{\nu}$, $E_{m}$, and $E_{r}$ generate visual feature $f^{\omega,\,i}_{t}$, $f^{\nu,\,i}_{t}$, $f_{r}$ and $f_{m}$, respectively.
\else
In MACMA framework (cf. section 3), the encoder $E_{\omega}$, $E_{\nu}$, $E_{m}$, and $E_{r}$ generate visual feature $f^{\omega,\,i}_{t}$, $f^{\nu,\,i}_{t}$, $f_{r}$ and $f_{m}$, respectively.
\EnableMainend
The disengagement quality of these learned features and states is important to the downstream policy learning or RIR generate.
In Figure~\ref{fig: visualization-learned-encoders}, we examine the semantics of visual features by overlaying the output of the visual encoder (from different layers) over the RGB images.
It is easy to see that the visual encoder has learned to pay more attention to the area (in red color) where the robot can walk. 
This effect becomes more evident as the encoder becomes deeper.
% --------------------------------------------------------------------
% \vspace{0.1in}
% --------------------------------------------------------------------

\subsection{On Dynamic Modality Importance for Action Selection} \label{appx: vap-action}
\begin{figure*}[t!]
\centering
\includegraphics[width=0.9\textwidth]{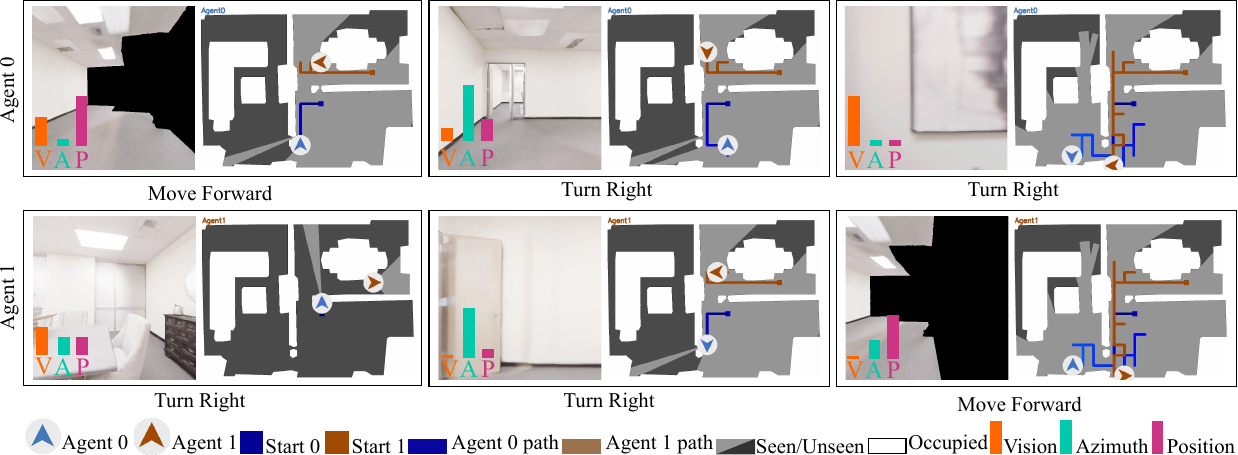}
\caption{\small 
	Dynamic vision, azimuth and position impact for two agents in an episode. 
	Columns correspond to three sampled time steps. The  orange, green and pink bars represent the importance of vision, azimuth and position, respectively.
}
\label{fig: impact-action}
\vspace{-6pt}
\end{figure*}

We postulate that at any given time step, the relative effects of visual (v), azimuth (a), and position (p) on agent decisions may vary.
To test our hypothesis, we 1) replace one of the above three modalities  (v, a, p) with random noise (such as v), 2) put the semi-corrupted input and the undamaged input in the trained model, and 3) get both the intervention action distribution and the original action distribution for a specific agent (e.g., agent 0), and 4) calculate the change, $d_{o, v}$ denotes the KL divergence of the original action distribution (o), and the intervened action distribution (v), which indicate the changes brought about by the intervention.
Then we do the same processes for the other modal inputs (e.g., a, p) in turn to get the change in azimuth intervention $d_{o, a}$ and changes in position intervention $d_{o, p}$.
$d_{o, a}$ denotes the KL divergence of the original action distribution (o) and the action distribution after the azimuth random intervention (a), $d_{o, p}$ denotes the KL divergence of the original action distribution (o) and the intervention action distribution (p).
Finally, normalize $d_{o, v}$, $\,d_{o, a}$ and $d_{o, p}$ to get $d^{*}_{o, v} $, $\,d^{*}_{o, a}$ and $^{*}d_{o, p}$.
$d^{*}_{o, v}$, $d^{*}_{o, a}$ and $^{*}d_{o, p}$ describe visual, azimuth, and position, respectively, which indicate the influence on the agent's decision-making~(\Eqref{sec: impact}).
Intuitively, the more drastic the normalized action changes, the more dependent the agent is on that modality.
Since there are two agents, the modality's influence value needs to be calculated separately for each agent.
The experimental intervention results for two agents are in Figure~\ref{fig: impact-action} (top for agent 0 and bottom for agent 1).

%------------------------------------------------------------------------------

%-------------------------------------------------------------------------------
\begin{equation} \label{sec: impact}
    \begin{split}
        & d^{*}_{o, v} = \nicefrac{d_{o, v}}{( d_{o, v} + d_{o, a} + d_{o, p} )}, \quad
        d^{*}_{o, a} = \nicefrac{d_{o, a}}{( d_{o, v} + d_{o, a} + d_{o, p} )}, \quad \\
        & d^{*}_{o, p} = \nicefrac{d_{o, p}}{( d_{o, v} + d_{o, a} + d_{o, p} )},
    \end{split}
\end{equation}

\subsection{On Dynamic Modality Importance for the RIR Measurement} \label{appx: vap-rir}

We hypothesized that the role of visual (v), azimuth (a), and position (p) in predicting the room impulse response might vary at any given time step.
To test our hypothesis, 1) we replace one modality (e.g., v) of the input (v, a, p) with random noise, 2) test this intervention and non-intervention input using the trained model for 1000 episodes, and 3) record the results of both the intervention test result PE(v) and test result PE without intervention, 4) calculate the change, we use $\delta^{v}_{PE}$ to denote the absolute change in test metric PE after visual input intervention (v).
By the same steps, we can get $\delta^{a}_{PE}$ (denote the absolute change in test metric PE after azimuth input intervention), and $\delta^{p}_{PE}$ (indicate the total shift in test metric PE after Position input intervention).
%------------------------------------------------------------------------------------------------------
\begin{equation} \label{exp-rnd-intervene}
	\begin{split}
		& \delta^{v}_{PE} = | PE - PE(v) |, \quad\quad \delta^{a}_{PE} = | PE - PE(a) |, \quad\quad  \\
		& \delta^{p}_{PE} = | PE - PE(p) |, \quad\quad  \delta^{v\star}_{PE} = \nicefrac{\delta^{v}_{PE}}{( \delta^{v}_{PE} + \delta^{a}_{PE} + \delta^{p}_{PE} )}, \\
		& \delta^{a\star}_{PE} = \nicefrac{\delta^{a}_{PE}}{( \delta^{v}_{PE} + \delta^{a}_{PE} + \delta^{p}_{PE} )}, \quad\quad
		\delta^{p\star}_{PE} = \nicefrac{\delta^{p}_{PE}}{( \delta^{v}_{PE} + \delta^{a}_{PE} + \delta^{p}_{PE} )}, 
	\end{split}
\end{equation}
% --------------------------------------------------------------------
Finally, normalize $\delta^{v}_{PE}$, $\delta^{a}_{PE}$, and $\delta^{p}_{PE}$ to get $\delta ^{v\star}_{PE}$, $\delta^{a\star}_{PE}$, and $\delta^{p\star}_{PE}$.
$\delta^{v\star}_{PE}$, $\delta^{a\star}_{PE}$, and $\delta^{p\star}_{PE}$ characterizes the impact of vision, azimuth and position on RIR prediction respectively in~\Eqref{exp-rnd-intervene}.
Intuitively, the larger the value of $\delta^{v\star}_{PE}$, $\delta^{a\star}_{PE}$, and $\delta^{p\star}_{PE}$, the more dependent the predicted RIR is on that modality.
Since there are two agents, the modality's influence value needs to be calculated separately for each agent.
The experimental intervention results for two agents are in Figure~\ref{fig: impact-action} (Top for agent 0 and bottom for agent 1).
Intuitively, the more significant the normalized damage degree, the greater the influence of this mode.
Since there are two agents, the influence value of the modality needs to be calculated separately for each agent.
% The experimental intervention results for two agents are in Table~\ref{tab: intervene-rnd-metrics}.
We test on dataset Replica under the settings: $\alpha^{\xi}$=1.0, $\alpha^{\zeta}$=1.0, $\alpha^{\psi}$=-1.0, $\alpha^{\phi}$=1.0, $\lambda$=0.1, $\kappa$=2, $\rho$=-1.0.
We can draw that the position input has the most significant impact on the prediction accuracy of RIR from Table~\ref{tab: intervene-rnd-metrics}.
% --------------------------------------------------------------------
\begin{table}[!ht]
    \centering
    \resizebox{0.95\linewidth}{!}{
        \begin{tabular}{ccc|ccc:ccc}
            \hline
            \multicolumn{3}{c|}{Random intervene} &  \multicolumn{3}{c:}{agent 0 ($\%$)}  & \multicolumn{3}{c}{agent 1 ($\%$) }      \\
            \cline{1-9}
            Vision & Azimuth & Position         & $\delta_{PE}$   & $\delta_{WCR}$  & $\delta_{RTE}$ & $\delta_{PE}$   & $\delta_{WCR}$  & $\delta_{RTE}$  \\
            \hline
            \Checkmark    & \XSolidBrush & \XSolidBrush  &   0.14     &  0.38  &  0.19 &   0.11     &  0.33  &  0.41    \\
            \XSolidBrush  & \Checkmark   & \XSolidBrush  &  16.53     & 19.28  & 47.92 &  33.64     & 28.24  &  9.80    \\
            \XSolidBrush  & \XSolidBrush & \Checkmark    &  83.33     & 80.34  & 51.89 &  66.25     & 71.43  & 89.79    \\
            \hline
        \end{tabular}
    }
    \caption{Dynamic importance of modality for the measure.}
    \label{tab: intervene-rnd-metrics}
\end{table}
% --------------------------------------------------------------------

\subsection{Visualization of Agents’ State Features}\label{appx: state-action}
\begin{figure}[h!]
    \centering
    \includegraphics[width=0.48\textwidth]{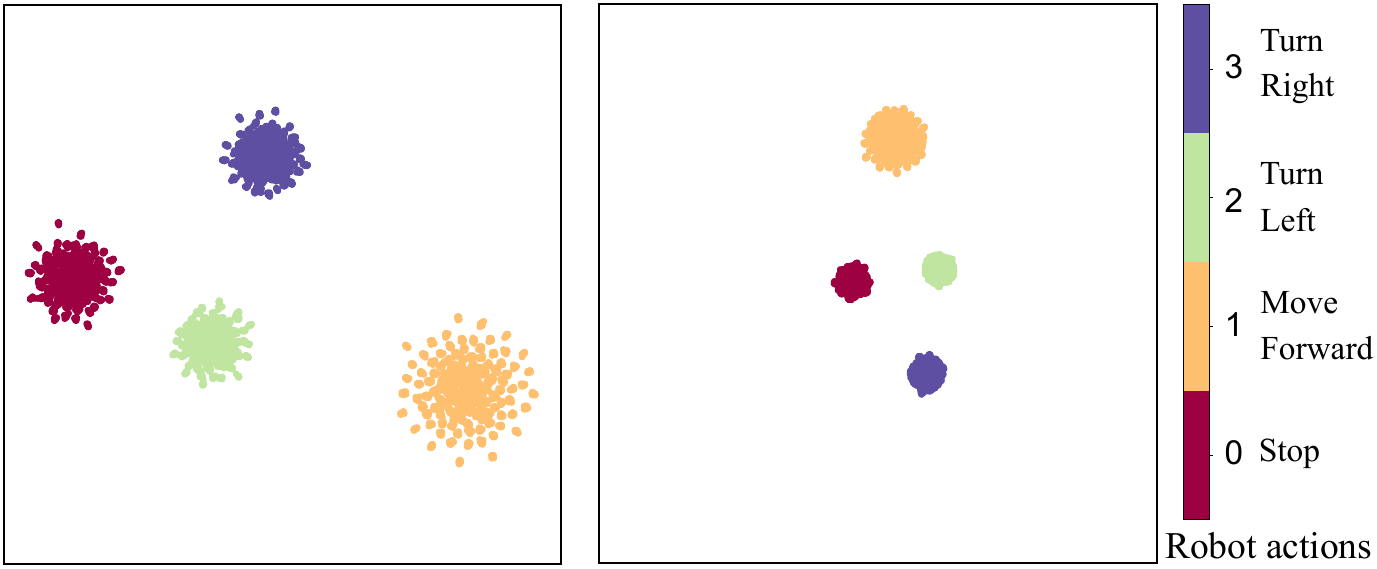}
    \caption{\small UMAP of  agent 0's state features (left) and agent 1's state features (right).
    }
    \label{fig: state-features}
    % \vspace{-5pt}
\end{figure}

It is more challenging to visualize the disengagement quality of the state representations $s^{\omega}_{t}$ and $s^{\nu}_{t}$. As a result, we choose to perform dimension reduction (to two dimensions) and clustering using UMAP~\cite{UMAP}.
We show the UMAP result in Figure~\ref{fig: state-features} with a color coding representing the action selected by the robot.
%------------------------------------------------------------------
The learned state representations correlate naturally with the robot's action selection in Figure~\ref{fig: state-features}.

% --------------------------------------------------------------------

\subsection{Algorithm Parameters}\label{appx: parameters}
The parameters used in our model are specified in Table ~\ref{tab: parameters}.
%---------------------------------------------------------------
\begin{table}[htb]
\vspace{-0.1in}
\centering
\resizebox{\linewidth}{!}{
\begin{tabular}{lcc} 
\hline
Parameter                & Replica(Matterport3D)    \\ 
\hline
number of updates       & 40000(60000)       \\
use linear learning rate  decay   & False     \\
use linear clip decay & False    \\
RIR sampling rate      & 16000   \\
clip param              & 0.1        \\
ppo epoch               & 4          \\
num mini batch         & 1         \\
value loss coef        & 0.5       \\
entropy coef            & 0.02     \\
learning rate        & $2 \times 10^{-4}$   \\
max grad norm          & 0.5      \\
num steps               & 150       \\
use GAE (Generalized Advantage Estimation)                  & True     \\
reward window size     & 50        \\
window length             & 600                       \\
window type               & ``hann window"  \\
number of processes     & 5(10)         \\
$w^{mse}$                & 1.0                      \\
$w^{\omega}_{m}, w^{\nu}_{m}, w_{m}, w_{\xi}$           & \nicefrac{1}{2}             \\   
$\alpha^{\xi}, \alpha^{\zeta}, \alpha^{\phi}$             &   1.0             \\
$\alpha^{\psi}$             &   -1.0            \\
$\beta$                      & 0.01     \\
$\gamma$                    & 0.99       \\
$\tau$                      & 0.95       \\
$\kappa$                       &    1               \\
$\lambda$                           &    0.1           \\
fft size                  & 1024                \\
shift size                & 120                    \\
hidden size             & 512         \\
optimizer               & Adam       \\
\hline
\end{tabular}
}
\caption{Algorithm parameters.}
\label{tab: parameters}
\end{table}

%##########################################################################################
% \section{MACMARA: MACMA WITH REWARD ASSIGNMENT}\label{appx: m2mrirra}
\section{MACMARA: MACMA with Reward Assignment}\label{appx: m2mrirra}

We present the extension MACMARA (Measuring Acoustics with Collaborative Multiple Agents with a dynamic Reward Assignment) here.
MACMARA is the MACMA add a learnable Reward Assignment module for environmental reward.
\subsection{The Overview of MACMARA}\label{appx: overview-m2mrirra}
\begin{figure*}[t!]
    \centering
    \includegraphics[width=\textwidth]{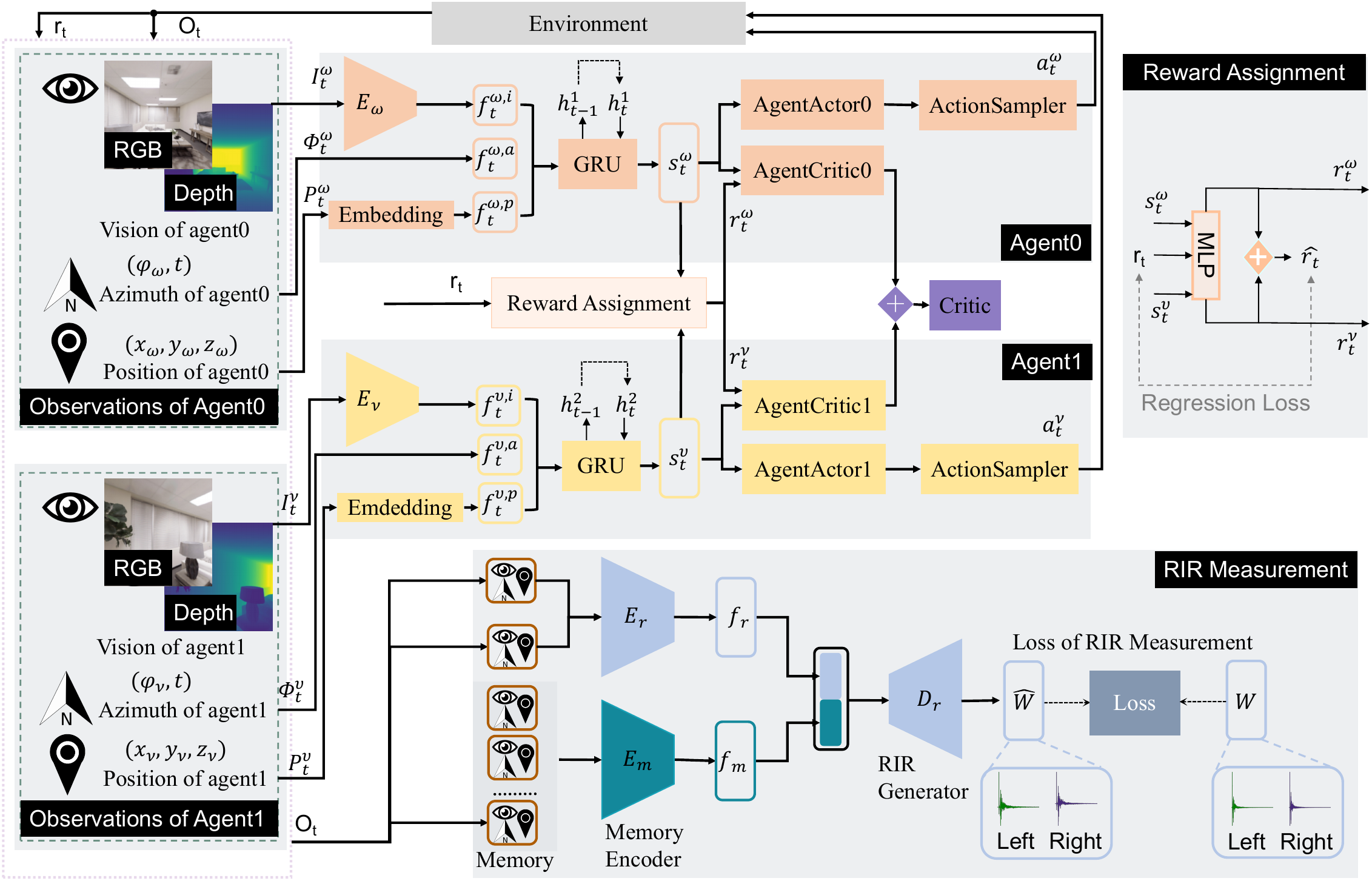}
    \caption{
        The MACMARA architecture: the agent 0 and the agent 1 first learn to encode observations as $s^{\omega}_{t}$ and $s^{\nu}_{t}$ respectively using encoder $E_{\omega}$ and $E_{\nu}$, which are fed to actor-critic networks to predict the next action $a_t^{\omega}$ and $a_t^{\nu}$.
        The reward assignment module is responsible for decomposing the reward into the agent 0's component $r_t^{\omega}$ and the agent 1's component $r_t^{\nu}$.
        The RIR Measurement learns how to predict room impulse response $\hat{W}_{t}$ guided by ground truth $W_{t}$.
        The reward assignment learns how to assign reward $r_{t}$ given by the environment supervised by $r_{t}$.
    }
    \label{fig: network-m2mrirra}
\end{figure*}

We also explored a model with with reward assignment.
This model has two cooperating agents moving in a 3D environment, using vision, position and azimuth to measure the RIR.
The proposed model mainly consists of four parts: agent 0, agent 1, RIR measurement  and environmental reward assignment(see Figure~\ref{fig: network-m2mrirra}).
Given egocentric vision, azimuth, and position inputs, our model encodes these multi-modal cues to 1) determine the action for agents and evaluate the action taken by the agents for policy optimization, 2) measure the room impulse response and evaluate the regression accuracy for the RIR generator, and 3) evaluate the trade off between the agents' exploration and RIR measurement,
4) environmental reward assignment. 
The two agents repeat this process until the maximal steps has been reached.

\EnableMainstart
As 1), 2), and 3) introduced in section~\ref{sec: approach} of the main paper, we mainly introduced the environmental reward assignment module here.
\else
As 1), 2), and 3) introduced in section 3 of the main paper, we mainly introduced the environmental reward assignment module here.
\EnableMainend
%----------------------------------------------------------------------------------------
\subsection{Environmental Reward Assignment for MACMARA} \label{appx: env-reward-assign-m2mrirra}
Here we focus on a learnable reward assignment module for environmental reward in MACMARA.
%-------------------------------------------------
% \setkeys{Gin}{draft} % make the graph not be seen, but make an placeholder
% \setkeys{Gin}{draft=false} % make the graph be seen, but make an placeholder
%----------------------------------------------------------------------------------------

\noindent{\textbf{Environmental reward assignment.}}\label{appx: m2mrirra-reward-assign}
Environmental reward assignment is implemented by a multilayer perceptron.
Given $s^{\omega}_{t}$, $s^{\nu}_{t}$ and $r_{t}$ at time step $t$, this module output $r^{\omega}_{t}$ and $r^{\nu}_{t}$.
To enforce this learning follow a desired direction, we add $r^{\omega}_{t}$ and $r^{\nu}_{t}$ to get $\hat{r}_{t}$.
Reward assignment is learned with the guide signal $r_{t}$.
The optimization objective of reward assignment is making $\hat{r}_{t}$ close to $r_{t}$.
We define a regression loss $\mathcal{L}^{\sigma}$ to perform this optimization:
\begin{equation} \label{eq: loss-reward-assign}
    \mathcal{L}^{\sigma} = \sum (r - {r}^{\omega} - {r}^{\nu})^{2},
\end{equation}
where ${r}^{\omega}$ and ${r}^{\nu}$ is respectively the predicted reward (for the $t$-th time step) of the agent 0 and agent 1.
${r}$ is the reward obtained from the environment.
$\mathcal{L}^{\sigma}$ is the regression loss for reward assignment module.
The distribution of rewards is a combination of trainable and fixed weighting:
\begin{equation} \label{eq: reward-ratio}
    \begin{split}
        &\rho^{\omega} = \frac{1 - \rho}{2} + \rho_{R}^{\omega} \cdot \rho, \quad
        \rho^{\nu} = \frac{1 - \rho}{2} + \rho_{R}^{\nu} \cdot \rho , \quad \\
        &\mathcal{L}^{\rho} = \sum (1 - \rho_{R}^{\omega} - \rho_{R}^{\nu})^{2}\ , \quad 0 \le \rho \le 1\ ,
    \end{split}
\end{equation}
where $\rho$ is a constant weight parameter;
$\rho_{R}^{\omega}$ and $\rho_{R}^{\nu}$ are reward weights predicted by neural networks;
$\rho^{\omega}$ and $\rho^{\nu}$ are immediate reward weights. $\mathcal{L}^{\rho}$ is equivalent to $\mathcal{L}^{\sigma}$ in \Eqref{eq: loss-reward-assign}.
Reward ${r}^{\omega}$ and ${r}^{\nu}$ can be respectively calculated with $r^{\omega} = r \cdot \rho^{\omega}$ and $r^{\nu} = r \cdot \rho^{\nu}$.

\subsection{The Loss of MACMARA} \label{appx: loss-m2mrirra}
The total loss of our model are formulated as \Eqref{eq: m2mrirra-loss-total}.
We minimize $\mathcal{L}$ following Proximal Policy Optimization (PPO)~\cite{ppo}.
\begin{equation} \label{eq: m2mrirra-loss-total}
    \mathcal{L} = w_{m} \cdot \mathcal{L}^{m} + w_{\xi} \cdot \mathcal{L}^{\xi} +  w_{\sigma} \cdot \mathcal{L}^{\sigma} ,
\end{equation}
where $\mathcal{L}^{m}$ is the loss component of motion for two agents,
$\mathcal{L}^{\xi}$ is the loss component of room impulse response prediction,
$\mathcal{L}^{\sigma}$ is the loss component of reward assignment for two agents,
$w_{m},\,w_{\sigma},\,w_{\xi}\,$ are hyperparameters.
\EnableMainstart
The loss of $\mathcal{L}^{m}$ are formulated as \Eqref{eq: loss-motion}.
\else
The loss of $\mathcal{L}^{m}$ are formulated as Equation 2.
\EnableMainend
\EnableMainstart
The loss of $\mathcal{L}^{\xi}$ are formulated as \Eqref{eq: loss-rir-prediction}.
\else
The loss of $\mathcal{L}^{\xi}$ are formulated as Equation 4.
\EnableMainend
The loss of $\mathcal{L}^{\sigma}$ are formulated as \Eqref{eq: loss-reward-assign}.
$w_{m}=w_{\xi}=w_{\sigma}=\nicefrac{1}{3}$.

\subsection{Formulation of MACMARA} \label{appx: formalation-m2mrirra}
We denote the agent 0 and agent 1 with superscript $\omega$ and $\nu$, respectively. 
The game $\mathcal{M} = (\mathcal{S}, (\mathcal{A}^{\omega}, \mathcal{A}^{\nu}) , \mathcal{P},(\mathcal{R}^{\omega}, \mathcal{R}^{\nu}))$ consists of state set $\mathcal{S}$, action sets $\mathcal{A}^{\omega}\,$ and $\mathcal{A}^{\nu}\,$, and a joint state transition function $\mathcal{P}\,$: $\mathcal{S} \times \mathcal{A}^{\omega} \times \mathcal{A}^{\nu} \rightarrow \mathcal{S}$. 
The reward function $\mathcal{R}^{\omega}\,: \mathcal{S} \times \mathcal{A}^{\omega} \times \mathcal{A}^{\nu} \times \mathcal{S} \rightarrow \mathbb{R} \,$ for agent 0 and $\mathcal{R}^{\nu} : \mathcal{S} \times \mathcal{A}^{\omega} \times \mathcal{A}^{\nu} \times \mathcal{S} \rightarrow \mathbb{R} \,$ for agent 1 respectively depends on the current state, next state and both the agent 0's and the agent 1's actions. 
Each player wishes to maximize their discounted sum of rewards. 
$r$ is the reward given by the environment at every time step in an episode.
MACMARA is modeled as a multi-agent~\cite{mavdn,maqmix} problem involving two collaborating players sharing the same goal:
% \vspace{-0.2in}
\begin{equation} \label{eq: m2mrirra-total-ques}
    \footnotesize
    \begin{aligned}
        min. & \quad \mathcal{L} \\
        \text{s.t.} &  \quad \pi^{\star} = (\pi^{\star,\,\omega},\pi^{\star,\,\nu}) = \!\!\!\!
        \mathop{\arg\max}\limits_{\pi^{\omega} \in \Pi^{\omega},\,\pi^{\nu} \in \Pi^{\nu} } G(\pi^{\omega},\,\pi^{\nu},\,r) \\
        &  = \left(
        \mathop{\arg\max}\limits_{\pi^{\omega} \in \Pi^{\omega}} G(\pi^{\omega},\,r),\ 
        \mathop{\arg\max}\limits_{\pi^{\nu} \in \Pi^{\nu} } G(\pi^{\nu},\,r) 
        \right),  \\
        & \quad G(\pi^{\omega}\!,\pi^{\nu}\!,r)\!=\! w^{\omega}G(\pi^{\omega}\!,r) + w^{\nu}G(\pi^{\nu}\!,r), \\
        & \quad G(\pi^{\omega}\!,r)\!=\!\textstyle\sum_{t=0} \gamma^{t}r_{t}\rho^{\omega},
        \quad G(\pi^{\nu}\!,r)\!=\!\sum_{t=0} \gamma^{t}r_{t}\rho^{\nu},\\
        \quad & \quad  \rho^{\omega}\!\!= (1-\rho)/2+ \rho \cdot \rho_{R}^{\omega},  
        \quad \rho^{\nu}\!\!= (1-\rho)/2+ \rho \cdot \rho_{R}^{\nu}, \\
        \quad &  \quad  \rho_{R}^{\omega} + \rho_{R}^{\nu} = 1, \quad w^{\omega} > 0, \quad w^{\nu} > 0, \quad 0 \le \rho \le 1,  \\
        \quad & \quad 0 \le \rho_{R}^{\omega} \le 1, \quad 0 \le \rho_{R}^{\nu} \le 1, \\
        & \quad min.  \quad \mathcal{L}^{\xi}, \quad min.  \quad \mathcal{L}^{\sigma},
    \end{aligned}
\end{equation}
where $\mathcal{L}$ is defined in~\Eqref{eq: m2mrirra-loss-total}.
$G(\pi^{\omega},\,\pi^{\nu},\,r)$ is the expected joint rewards for agent 0 and agent 1 as a whole. 
$G(\pi^{\omega},\,r)$ and $G(\pi^{\nu},\,r)$ are the discounted and cumulative rewards for agent 0 and agent 1, respectively.
$w^{\omega}$ and $w^{\nu}$ denote the constant cumulative rewards balance factors for agent 0 and agent 1, respectively.
$\rho^{\omega}$ and $\rho^{\nu}$ are immediate reward contribution for agent 0 and agent 1, respectively.
$\rho_{R}^{\omega}$ and $\rho_{R}^{\nu}$ are present the trainable reward decomposition weights.
$\rho$ is a constant (throughout the training) reward allocation parameter.
\EnableMainstart
The loss of $\mathcal{L}^{\xi}$ are formulated as \Eqref{eq: loss-rir-prediction}.
\else
The loss of $\mathcal{L}^{\xi}$ are formulated as  Equation 4.
\EnableMainend
The loss of $\mathcal{L}^{\sigma}$ are formulated as \Eqref{eq: loss-reward-assign}.

\subsection{Experimental Results of MACMARA} \label{appx-exp-m2mrirra}
In experiments below, we use two models: MACMA and the extension MACMARA (Measuring Acoustics with Collaborative
Multiple Agents with a dynamic reward assignment module), to present the ablation on $\rho$.

$\rho$ is a constant reward weight (See \Eqref{eq: reward-ratio}).
Results are run on dataset Replica under the experimental settings: $\alpha^{\xi}$=1.0, $\alpha^{\zeta}$=1.0, $\alpha^{\psi}$=-1.0, $\alpha^{\phi}$=1.0, $\kappa$=2, $\lambda$=0.1.
As can be seen from Table ~\ref{tab: abla-rho} $\rho$=-1.0 is the best choice.

% ------------------------------------------
\begin{table}[h]
    \centering
    % \resizebox{0.95\linewidth}{!}{
    \begin{tabular}{c|ccccc}
        \hline
        $\rho$ & WCR ($\uparrow$) & PE ($\downarrow$)  & CR ($\uparrow$) & RTE ($\downarrow$)  & SiSDR ($\uparrow$)   \\
        \hline
            -1.0${}^{\heartsuit} $    & \textbf{0.6977} &  3.2509 & \textbf{0.7669} & 13.8896 & 23.6501  \\
        \hdashline
            0.0${}^{\clubsuit}$      & 0.6614 &  3.5727 & 0.7288 & 12.6987 & 23.1183  \\
            0.2     & 0.5287 &  3.3398 & 0.5798 & 14.0822 & 23.6715  \\
            0.4     & 0.5040 &  3.1860 & 0.5512 & 15.3278 & 23.6485  \\
            0.6     & 0.5384 &  3.3099 & 0.5904 & \textbf{12.4184} & 23.2581  \\
            0.8     & 0.4510 &  \textbf{2.9755} & 0.4903 & 14.5194 & \textbf{23.8725}  \\
            1.0     & 0.5471 &  3.2316 & 0.5995 & 12.9246 & 23.8201  \\
        \hline
    \end{tabular}
    % note
    \vspace{1px}
    
     {\raggedright
     \small
     ${\heartsuit}$: Without reward assignment module, and every agent get all the environment reward ($r^{\omega}_{t}=r^{\nu}_{t}=r_{t}$).
     
     \small
    ${\clubsuit}$: With reward assignment module, and every agent get all the environment reward ($r^{\omega}_{t}=r^{\nu}_{t}=0.5 r_{t}$).
    
     \par}
    % }
    \caption{Ablation on $\rho$.}
    \label{tab: abla-rho}
\end{table}
% --------------------------------------------------------------------

\subsection{Algorithm Parameters in MACMARA}\label{appx: m2mrirra-parameters}
The parameters used in our model MACMARA are specified in Table~\ref{tab: parameters} except $w_{m}=w_{\xi}=w_{\sigma}=\nicefrac{1}{3}$.

%%%%%%%%%%%%%%%%%%%%%%%%%%%%%%%%%%%%%%%%%%%%%%%%%%%%%%%%%%%%%%%%%%%%%%%%
\EnableAppxend  % 附录结束
%%%%%%%%%%%%%%%%%%%%%%%%%%%%%%%%%%%%%%%%%%%%%%%%%%%%%%%%%%%%%%%%%%%%%%%%
%####################################################

% \appendix

\bibliographystyle{named}
\bibliography{ijcai23}

\end{document}